\documentclass[lettersize,journal]{IEEEtran}
\usepackage{amsmath,amsfonts}
\usepackage{algorithmic}
\usepackage{algorithm}
\usepackage{array}
\usepackage{textcomp}
\usepackage{stfloats}
\usepackage{url}
\usepackage{verbatim}
\usepackage{graphicx}
\usepackage{cite}
\usepackage{booktabs}
\usepackage{xcolor}
\usepackage{multicol}
\usepackage{multirow}
\usepackage{amssymb}
\usepackage{placeins}
\usepackage{subfigure}
\begin{document}

\newcommand{\todo}[1]{\textcolor{red}{TODO: #1}}
\newcommand{\naishan}[1]{\textcolor{blue}{NaiShan: #1}}

\title{RexUniNLU: Recursive Method with Explicit Schema Instructor for Universal NLU}

\author{
\IEEEauthorblockN{Chengyuan Liu$^{1,2,\dag}$, Shihang Wang$^{2,\dag}$, Fubang Zhao$^{2}$, Kun Kuang$^{1,\ast}$,\\Yangyang Kang$^{2,\ast}$, Weiming Lu$^{1}$, Changlong Sun$^{2}$, Fei Wu$^{1}$}\\
\IEEEauthorblockA{$^1$ College of Computer Science and Technology, Zhejiang University}\\
\IEEEauthorblockA{$^2$ Institute for Intelligent Computing, Alibaba Group}\\
\IEEEauthorblockA{\small \texttt{\{liucy1, kunkuang, luwm, wufei\}@zju.edu.cn, \{wangshihang.wsh, fubang.zfb, yangyang.kangyy\}@alibaba-inc.com, changlong.scl@taobao.com}}
\thanks{$^{\dag}$ Equal contribution.}
\thanks{$^{\ast}$ Corresponding author.}
\thanks{This work was done when Chengyuan Liu interned
at Alibaba.}
}



\maketitle

\begin{abstract}
Information Extraction (IE) and Text Classification (CLS) serve as the fundamental pillars of NLU, with both disciplines relying on analyzing input sequences to categorize outputs into pre-established schemas. However, there is no existing encoder-based model that can unify IE and CLS tasks from this perspective. 
To fully explore the foundation shared within NLU tasks, we have proposed a \textbf{Recursive Method with Explicit Schema Instructor for Universal NLU}.
Specifically, we firstly redefine the true universal information extraction (UIE) with a formal formulation that covers almost all extraction schemas, including quadruples and quintuples which remain unsolved for previous UIE models. Then, we expands the formulation to all CLS and multi-modal NLU tasks.
Based on that, we introduce RexUniNLU, an universal NLU solution that employs explicit schema constraints for IE and CLS, which encompasses all IE and CLS tasks and prevent incorrect connections between schema and input sequence. To avoid interference between different schemas, we reset the position ids and attention mask matrices.
Extensive experiments are conducted on IE, CLS in both English and Chinese, and multi-modality, revealing the effectiveness and superiority.
Our codes are publicly released\footnote{https://modelscope.cn/models/damo/nlp\_deberta\_rex-uninlu\_chinese-base/summary}.
\end{abstract}


\section{Introduction}

\IEEEPARstart{T}{he} advent of large language models (LLMs) \cite{touvron2023llama} has ushered in a new era of unifying natural language processing tasks, particularly revolutionizing the domain of natural language generation (NLG). However, within the realm of Natural Language Understanding (NLU)—where precise output concerning labels or spans is paramount for tasks like Information Extraction (IE) and Text Classification (CLS)—Large Language Models (LLMs) often fall short of expectations. Their nature of generation can lead to the prediction of erroneous category labels or the identification of entity spans that don't actually exist in the input sequence. Accordingly, the LLMs also struggle to precisely pinpoint the requested positions of entity spans in the input sequence. Moreover, the decoder-based generation architecture of the LLMs are frequently criticized for their low efficiency. Consequently, the significance of deploying encoder based small language models within the NLU tasks should still be addressed.

\begin{figure}[t]
    \centering
    \includegraphics[width=\linewidth]{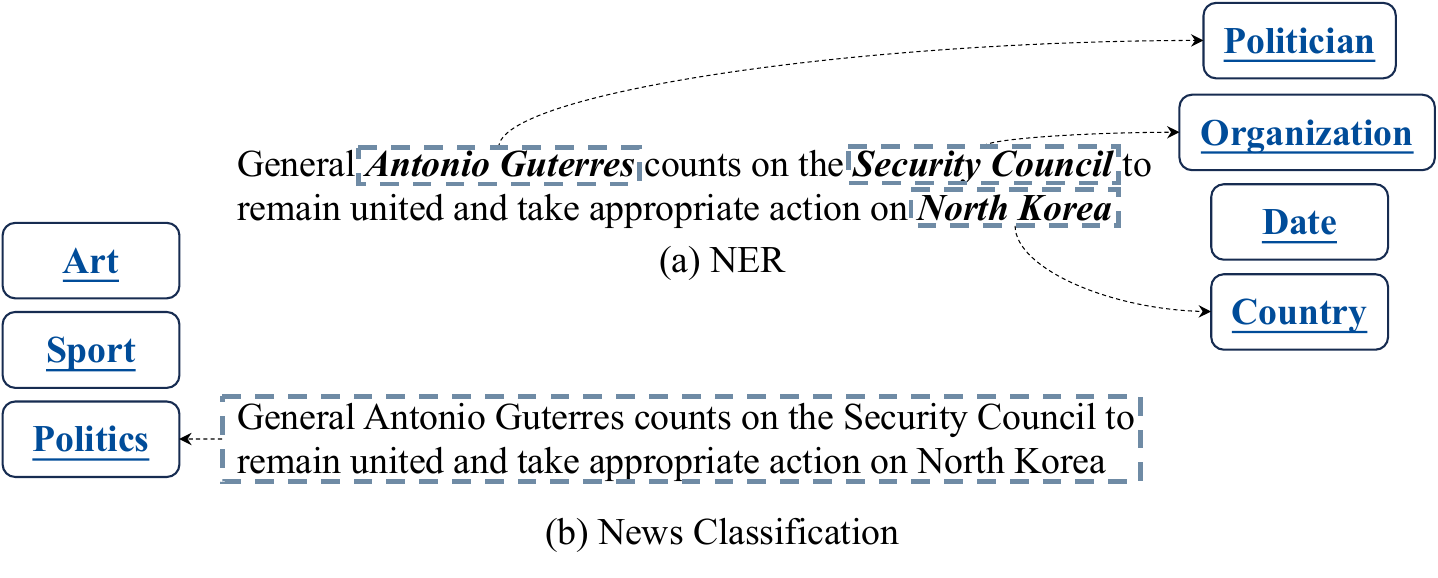}
    \caption{Similarity between IE and CLS. IE focuses on establishing the relationship between pre-defined labels and specific components of the input sequence, whereas CLS aims to connect these labels with the entirety of the input sequence.}
    \label{fig:clsandie}
\end{figure}

Although IE
and CLS
are generally considered as the two backbone tasks within NLU, they are studied separately by current community, hindering the fact that IE and CLS tasks are both establishing the relationship between pre-defined labels and the input sequence, as shown in Figure \ref{fig:clsandie}.

Some studies attempted to handle NER \cite{tkde:ner1}, RE \cite{tkde:re1,tkde:re2,tkde:re3}, EE \cite{tkde:eventdetection1} and ABSA \cite{tkde:absa} with universal information extraction (UIE) framework.
T5-UIE \cite{duuie} designed novel Structural Schema Instructor (SSI) as inputs and Structured Extraction Language (SEL) as outputs, and proposed a unified text-to-structure generation framework based on T5-Large. While USM \cite{usm} introduced three unified token linking operations and uniformly extracted substructures in parallel, which achieves new SoTAs on IE tasks.


\begin{figure}[t]
    \centering
    \includegraphics[width=\linewidth]{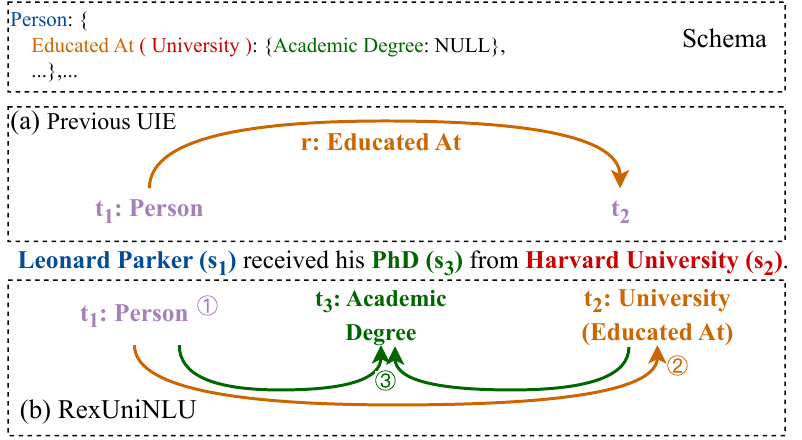}
    \caption{Comparison of RexUniNLU with previous UIE. (a) The previous UIE models the information extraction task by defining the text spans and the relation between span pairs, but it is limited to extracting only two spans. (b) Our proposed RexUniNLU recursively extracts text spans for each type based on a given schema, and feeds the extracted information to the following extraction.}
    \label{fig:redefine_uie}
\end{figure}

However, they fall short of being true UIE models particularly when extracting other general schemas such as quadruples and quintuples. Additionally, these models used an implicit structural schema instructor, which could lead to incorrect links between types, hindering the model's generalization and performance in low-resource scenarios.
As illustrated in Figure \ref{fig:redefine_uie} (a), previous UIE can only extract a pair of spans along with the relation between them, while ignoring other qualifying spans (such as location, time, etc.) that contain information related to the two entities and their relation.
Moreover, previous UIE models are short of explicitly utilizing extraction schema to restrict outcomes. The relation \textit{work for} provides a case wherein the subject and object are the \textit{person} and \textit{organization} entities, respectively. Omitting an explicit schema can lead to spurious results, hindering the model's generalization and performance in resource-limited scenarios.

There are also some studies who unified CLS tasks, such as Natural Language Inference (NLI) and Sentiment Analysis, while \textbf{they cannot handle IE}. UniMC \cite{unimc} proposed a new paradigm for zero-shot learners. It is compatible with any format and applicable to a list of language tasks, such as text classification, commonsense reasoning, coreference resolution, and sentiment analysis. UTC \cite{utc} (Unified text classification) is another successful work. UTC models the relationship between labels and texts with Unified Semantic Matching, supporting NLU tasks under low-resource settings, such as general text classification, sentiment analysis, natural language inference, multi-choice machine reading comprehension. One of the key weakness of these studies is that they are unable to extract explicit spans according to the given schema. SiameseUniNLU \cite{siameseuninlu} unifies several NLU tasks, including IE and CLS, within a framework, but it cannot handle hierarchical labels. Besides, the efficiency of SiameseUniNLU is poor as it can process only one pre-established schema item at one time when handling IE.
Moreover, generative based approaches also contend with the same issues of low efficiency and lack of control that plague the LLMs \cite{duuie}.

To fully explore the advantage of the shared pattern within NLU tasks, we have introduced a \textbf{Recursive Method with Explicit Schema Instructor for Universal NLU} in this paper, which incorporates all NLU tasks in a unified pipeline.
We firstly redefine the true UIE with a formal formulation that covers almost all extraction and classification schemas, and the formulation also expands to all text classification and multi-modal NLU tasks. To the best of our knowledge, we are the first to unify UIE and other classification task covering all extraction and classification schemas.

Additionally, we propose RexUniNLU, which incorporates the NLU tasks covering all schemas, using explicit class constraints for IE and CLS.
RexUniNLU recursively runs queries for all schema types and utilizes three unified token-linking operations to compute the results of each query.
We construct an Explicit Schema Instructor (ESI), providing rich label semantic information to RexUniNLU, and assuring the extraction and classification results meet the constraints of the schema. ESI and the text are concatenated to form the query.

Take Figure \ref{fig:redefine_uie} (b) as an example, given the extraction schema, RexUniNLU firstly extracts ``Leonard Parker'' classified as a ``Person'', then extracts ``Harvard University'' classified as ``University'' coupled with the relation ``Educated At'' according to the schema. Thirdly, based on the extracted tuples ( \textit{``Leonard Parker'', ``Person''} ) and ( \textit{``Harvard University'', ``Educated At (University)''} ), RexUniNLU derives the span ``PhD'' classified as an ``Academic Degree''.
RexUniNLU extracts spans recursively based on the schema, allowing extracting more than two spans such as quadruples and quintuples, rather than exclusively limited to pairs of spans and their relation.
To avoid interference between different classes, we reset the position ids and attention mask matrices.

Extensive experiments are conducted on information extraction, text classification in both Chinese and English, and multi-modality, revealing the effectiveness and superiority.
We pre-trained RexUniNLU on a combination of supervised IE datasets, Machine Reading Comprehension (MRC) datasets, text classification datasets, as well as over 9 million Joint Entity and Relation Extraction (JERE) instances constructed via Distant Supervision. Apart from single modality tasks, we have also expanded the capabilities of the RexUniNLU model to multi-modal settings, namely MRexUniNLU. All the experiments demonstrate that RexUniNLU and MRexUniNLU surpasses the state-of-the-art performance in various tasks and outperforms previous UIE and NLU models in low-resource and high-resource settings. The RexUniNLU large and base models respectively claim the top two spots on the PCLUE leaderboard \footnote{https://www.cluebenchmarks.com/pclue.html}. Additionally, RexUniNLU exhibits remarkable superiority in handling complex tasks such as extracting quadruples and quintuples.

Our contributions can be summarized in 4 folds:
\begin{enumerate}
    \item We redefine the true UIE with a formulation that covers almost all extraction schema.
    \item The formulation also expands to all NLU tasks. Thus IE and CLS can be solved in a unified manner.
    \item We introduce RexUniNLU, which recursively runs queries for all schema types and utilizes three unified token-linking operations to compute the outcomes of each query. It employs explicit schema instructions to augment label semantic information and enhance the performance in low and high resource scenarios.
    \item We pretrain RexUniNLU to improve the performance in low-resources scenarios. RexUniNLU achieved new SoTAs on several datasets. Extensive experiments are conducted on information extraction, text classification, and multi-modality, revealed the effectiveness and superiority.
\end{enumerate}

\section{Related Work}\label{sec:related work}

\subsection{Information Extraction}

Task-specific models for IE have been extensively studied, including Named Entity Recognition \cite{wang-etal-2021-improving}, Relation Extraction \cite{RE1}, Event Extraction \cite{DBLP:journals/corr/abs-2108-10038}, and Aspect-Based Sentiment Analysis \cite{zhang-etal-2021-towards-generative, xu-etal-2021-learning}.

Some recent works attempted to jointly extract the entities, relations and events.
T5-UIE \cite{duuie} is the unified structure generation for UIE. 
They proposed a framework based on T5 architecture to generate SEL containing specified types and spans. However, the auto-regressive method suffers from uncontrollability and low GPU utilization. USM \cite{usm} is an end-to-end framework for UIE, by designing three unified token linking operations. Empirical evaluation on 4 IE tasks showed that USM has strong generalization ability in zero/few-shot transfer settings.
FSUIE \cite{peng-etal-2023-fsuie} is enhanced with fuzzy span loss and fuzzy span attention.
UniEx \cite{ping-etal-2023-uniex} converts the text-based IE tasks as the token-pair problem.

\subsection{Text Classification}

Given an input sentence and optional contexts, text classification (CLS) aims to classify the sentence with one of the pre-defined labels. CLS covers several classic NLU tasks such as Natural Language Inference (NLI) \cite{N18-1101,snliemnlp2015}, Sentiment Analysis \cite{maas-EtAl2011ACL-HLT2011,10.1145/2507157.2507163,marc_reviews} and Multi-choice MRC.

As methods have advanced, more and more attention has been paid to how to unify these tasks. UniMC \cite{unimc} is proposed for zero-shot learners. It is compatible with any format and applicable to a list of language tasks, such as text classification, commonsense reasoning, coreference resolution, and sentiment analysis. UniMC shows state-of-the-art performance on several benchmarks and produces satisfactory results on tasks such as NLI. UTC is developed based on USM \cite{usm}. UTC models the relationship between labels and texts with Unified Semantic Matching, supporting NLU tasks under low-resource settings, such as general text classification, sentiment analysis, NLI, multi-choice machine reading comprehension. SiameseUniNLU \cite{wang2019structbert, Zhao2021AdjacencyLO} supports general text classification and IE based on Siamese neural network. The cross-attention in SiameseUniNLU is only applied to the top-n layers, leading to higher efficiency.

\subsection{Zero-Shot and Few-Shot}

Low-resources setting is more practical than full-shot. For few-shot, the model is trained on several samples with each category. While for zero-shot \cite{tkde:ZhengCWZYZ24}, no training data is available, which is a challenging setting.
Pre-training \cite{tkde:lm1,tkde:lm2,tkde:autoalign} is a powerful strategy to improve the performance on low-resources downstream tasks.
GPT-2 \cite{radford2019language} is pre-trained in the manner of self-regression to handle few-shot problems.
GPT-4 extends to multi-modality, GPT-4 can accept image and text inputs and produce text outputs. Llama \cite{touvron2023llama} is a collection of foundation language models ranging from 7B to 65B parameters, trainable with several GPUs. 
However, due to the differences in task forms, LLMs are relatively poor especially at extracting the offsets of spans or precisely generating the original content from the input sequence. Additionally, it is a great challenge for privatization deployment and fine-tuning, because of the high cost on computation and memory.

Apart from LLMs, several existing pre-trained generative language models that are capable of handling various few-shot or zero-shot NLU tasks \cite{duuie,promptclue,zsac}. Nonetheless, they encounter the same nature of unpredictability as LLMs, which, at times, can be exacerbated owing to their smaller scale. Furthermore, the development of encoder-based small language models for universal NLU tasks under low-resource settings, remains largely unexplored. 

\subsection{Cross Modality NLU}

LayoutLM \cite{xu2020layoutlm} first leveraged both text and layout information across scanned document. Subsequent developments incorporated image features to enhance performance in IE and CLS tasks \cite{huang2022layoutlmv3}. However, it is commonly acknowledged that image features were the bottleneck of model efficiency and they could be superfluous in text-dense documents like contracts, notifications, and legal judgements, etc. Nonetheless, the majority of the previous works were pre-trained with fixed modality combination, limiting their flexibility in the downstream applications.

Furthermore, the scarcity of high-quality annotated data across a wide array of multi-modal tasks presented a considerable challenge, leading to limited research on document understanding capabilities in few-shot and zero-shot contexts.
Current studies \cite{
luo2024layoutllm} endeavored to address this shortfall by leveraging the LLMs, but as previously noted, LLMs grapple with intrinsic constraints when tackling NLU tasks. Besides, deploying a LLM could be quite expensive.

\section{Universal Natural Language Understanding}

We start from redefining universal information extraction with a formulation in SubSection \ref{sec:redefine uie}. Then the formulation expands to general text classification tasks.

\subsection{Redefine UIE}\label{sec:redefine uie}

While previous studies \cite{duuie, usm} proposed Universal Information Extraction as methods of addressing NER, RE, EE, and ABSA with a single unified model, their approaches were limited to only a few tasks and ignored schemas that contain more than two spans, such as quadruples and quintuples. Hence, we redefine UIE to cover the extraction of more general schemas.

In our view, genuine UIE extracts a collection of structured information from the text, with each item consisting of $n$ spans $\mathbf{s}=[s_1, s_2, \dots, s_n]$ and $n$ corresponding types $\mathbf{t}=[t_1, t_2, \dots, t_n]$. The spans are extracted from the text, while the types are defined by a given schema. Each pair of $(s_i, t_i)$ is the target to be extracted.

Formally, we propose to maximize the probability in Equation \ref{new UIE}.
\begin{equation}
\label{new UIE}
\begin{aligned}
  &\prod_{ (\mathbf{s}, \mathbf{t} ) \in \mathbb{A} } p \big ( ( \mathbf{s}, \mathbf{t}  ) | \mathbf{C}^n, \mathbf{x} \big ) \\
  = &\prod_{ (\mathbf{s}, \mathbf{t}) \in \mathbb{A} } \prod_{i=1}^n p\left( \left(s, t \right )_i | \left (\mathbf{s}, \mathbf{t} \right )_{<i}, \mathbf{C}^n, \mathbf{x} \right ) \\
  = &\prod_{i=1}^n \left [ \prod_{ \left (s, t \right )_i \in \mathbb{A}_i | \left (\mathbf{s}, \mathbf{t} \right )_{< i} } p\left ( \left (s, t \right)_i | \left ( \mathbf{s}, \mathbf{t} \right )_{<i}, \mathbf{C}^n, \mathbf{x} \right) \right ]
\end{aligned}
\end{equation}
where $\mathbf{C}^n$ denotes the hierarchical schema (a tree structure) with depth $n$, $\mathbb{A}$ is the set of all sequences of annotated information. $\mathbf{t}=[t_1, t_2, \dots, t_n]$ is one of the type sequences (paths in the schema tree), and $\mathbf{x}$ is the text. $\mathbf{s}=[s_1, s_2, \dots, s_n]$ denotes the corresponding sequence of spans to $\mathbf{t}$.
We use $(s, t)_i$ to denote the pair of $s_i$ and $t_i$. Similarly, $\left ( \mathbf{s}, \mathbf{t} \right )_{<i}$ denotes $[s_1, s_2, \dots, s_{i-1}]$ and $[t_1, t_2, \dots, t_{i-1}]$. $\mathbb{A}_i | \left (\mathbf{s}, \mathbf{t} \right )_{<i}$ is the set of the $i$-th items of all sequences led by $\left (\mathbf{s}, \mathbf{t} \right )_{<i}$ in $\mathbb{A}$.
To more clearly clarify the symbols, we present some examples in Appendix \ref{sec:schema case}.

\subsection{Formalize UniNLU}

It is noteworthy that IE and CLS exhibit a consistent pattern where IE links pre-defined labels with particular segments of the input sequence, while CLS associates the labels with the entire input sequence. Consequently, extending the framework of UIE to encompass general CLS tasks and cover the full spectrum of NLU tasks appears to be a relatively straightforward endeavor. This extension could include tasks such as text classification (CLS), sentiment analysis (Sentiment), text matching (Match), natural language inference(NLI), and machine reading comprehension (MRC) in multiple-choice format, all of which can be categorized under the umbrella of generalized CLS tasks.

\begin{equation}\label{cls formulation}
    \prod_{i=1}^n \left [ \prod_{ t_i \in \mathbb{A}_i | \mathbf{t}_{< i} } p\left ( t_i | \mathbf{t}_{<i}, \mathbf{C}^n, \mathbf{x} \right) \right ]
\end{equation}

It is obvious that Equation \ref{new UIE} comprises Equation \ref{cls formulation}. To indicate the CLS task, we intuitively add a fixed \texttt{[CLST]} token ahead the input sequence, somewhat akin to the \texttt{[CLS]} token used in typical CLS tasks to represent the entire sequence. This \texttt{[CLST]} token functions as the extraction span, thereby generalizing the UIE framework within the context of NLU.

\section{RexUniNLU}

\begin{figure}[t]
    \centering
    \includegraphics[width=\linewidth]{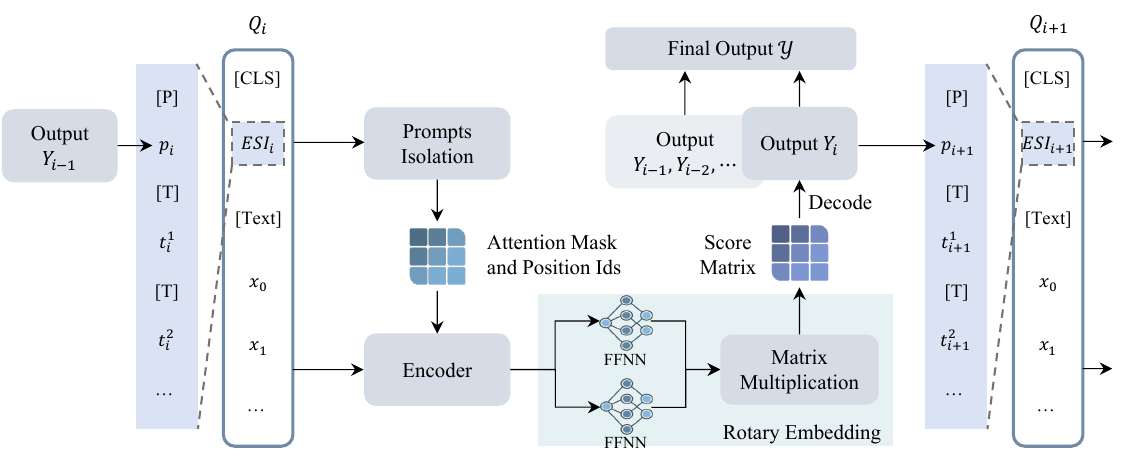}
    \caption{The overall framework of RexUniNLU. We illustrate the computation process of the $i$-th query and the construction of the $i+1$-th query. $Y_i$ denotes the output of the $i$-th query, with all outputs ultimately combined to form the overall extraction result.}
    \label{fig: framework}
\end{figure}

\begin{figure*}[t]
    \centering
    \subfigure[] {
        \centering
        \includegraphics[width=0.6\linewidth]{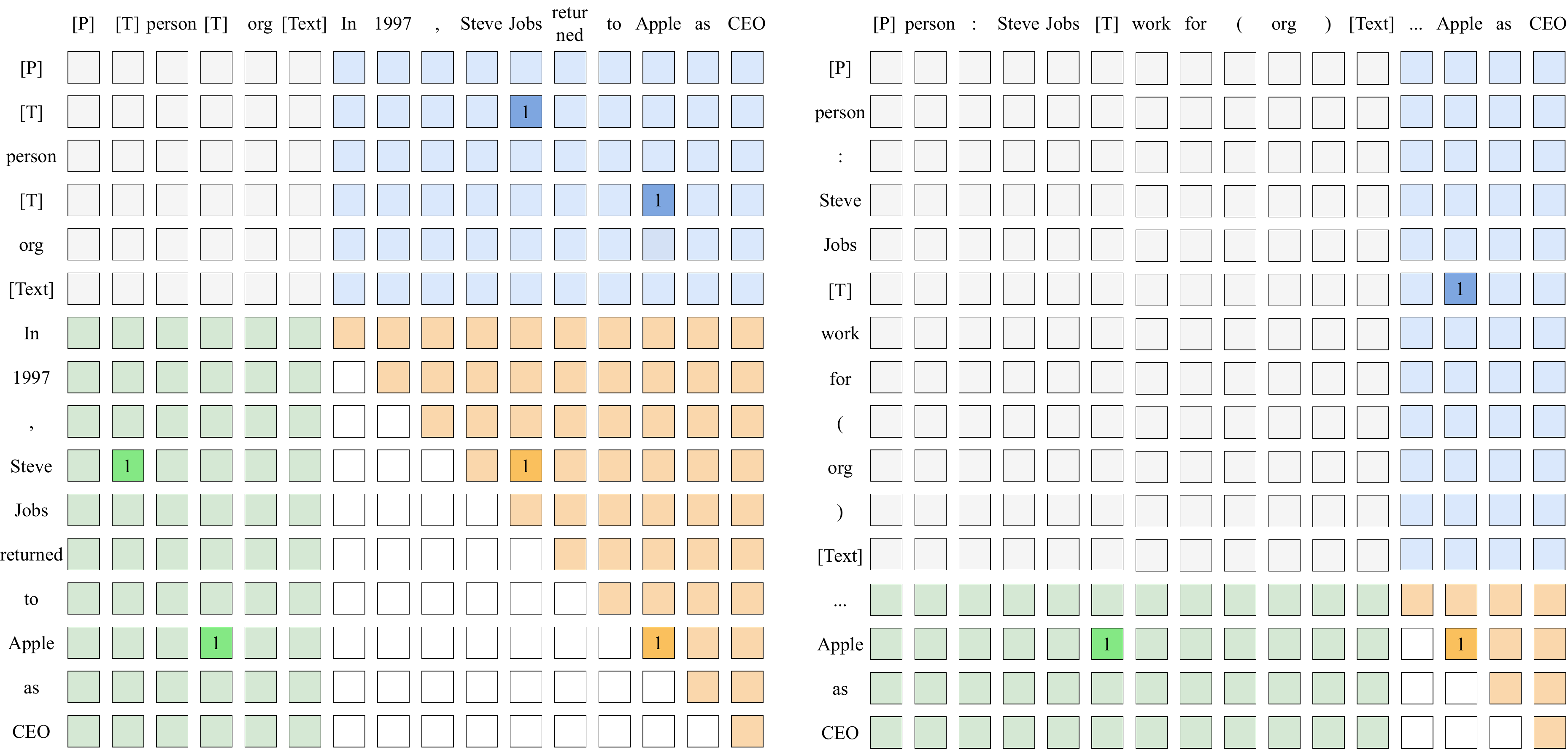}
        \label{fig: sample}
    }
    \subfigure[] {
        \centering
        \includegraphics[width=0.3\linewidth]{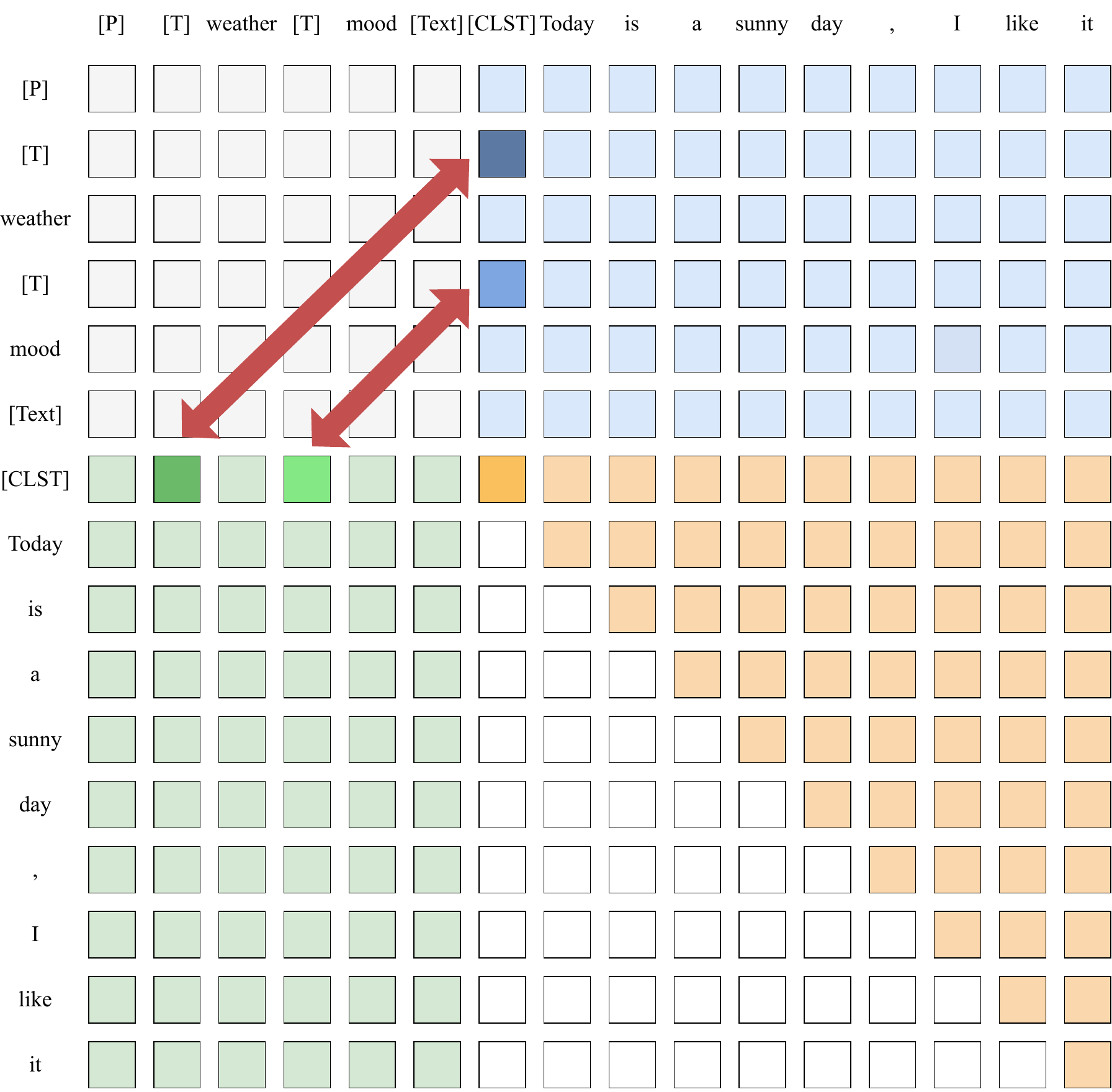}
        \label{fig:cls-metrix}
    }
    \caption{(a) Queries and score matrices for NER and RE. The left sub-figure shows how to extract entities ``Steve Jobs'' and ``Apple''. The right sub-figure shows how to extract the relation given the entity ``Steve Jobs'' coupled with type ``person''. The schema is organized as \textit{\{``person'': \{``work for (organization)'': null\}, ``organization'': null \}}. The score matrix is separated into three valid parts: \textcolor{orange}{token head-tail}, \textcolor{blue}{type-token tail} and \textcolor{green}{token head-type}. The cells scored as 1 are darken, the others are scored as 0. (b) The query and score matrix for text classification using RexUniNLU.}
\end{figure*}


We first introduce RexUniNLU under information extraction context, which models the learning objective Equation \ref{new UIE} as a series of recursive queries, with three unified token-linking operations employed to compute the outcomes of each query. The condition $(\mathbf{s, t})_{<i}$ in Equation \ref{new UIE} is represented by the prefix in the $i$-th query, and $(s, t)_i$ is calculated by the linking operations. Then we present RexUniNLU under classification tasks.


\subsection{RexUniNLU for IE}\label{sec:IE}

\subsubsection{Overall Framework}\label{sec:framework}
Figure \ref{fig: framework} shows the overall framework.
RexUniNLU recursively runs queries for all schema types. Given the $i$-th query $Q_i$, we adopt a pre-trained language model as the encoder to map the tokens to hidden representations $h_i \in \mathbb{R}^{n \times d}$, where $n$ is the length of the query, and $d$ is the dimension of the hidden states,
\begin{equation}
    h_i = \mathbf{Encoder}(Q_i, P_i, M_i)
\end{equation}
where $P_i$ and $M_i$ denote the position ids and attention mask matrix of $Q_i$ respectively..

Next, the hidden states are fed into two feed-forward neural networks $\mathbf{FFNN}_q, \mathbf{FFNN}_k$ .

Then we apply rotary embeddings following GlobalPointer \cite{su2022global} to calculate the score matrix $Z_i$.

\begin{equation}\label{equation: rotary}
    \begin{aligned}
    Z_i^{j,k} = ({\mathbf{FFNN}_q(h_i^j)}^{\top} \mathbf{R}(P_i^k-P_i^j) \\{\mathbf{FFNN}_k(h_i^k)}) \otimes M_i^{j, k}
    \end{aligned}
\end{equation}
where $M_i^{j, k}$ and $Z_i^{j,k}$ denote the mask value and score from token $j$ to $k$ respectively. $P_i^j$ and $P_i^k$ denote the position ids of token $j$ and $k$. $\otimes$ is the Hadamard product. $\mathbf{R}(P_i^k-P_i^j) \in \mathbb{R}^{d\times d}$ denotes the rotary position embeddings (RoPE), which is a relative position encoding method with promising theoretical properties.

Finally, we decode the score matrix $Z_i$ to obtain the output $Y_i$, and utilize it to create the subsequent query $Q_{i+1}$. All ultimate outputs are merged into the result set $\mathcal{Y} = \{ Y_1, Y_2, \dots  \}$.

We utilize Circle Loss \cite{su2022global} as the loss function of RexUniNLU, which is very effective in calculating the loss of sparse matrices
\begin{equation}
\begin{aligned}
    \mathcal{L}_i = \log (1 + \sum_{\hat{Z}_i^j = 0} e^{\overline{Z}_i^j})& + \log (1 + \sum_{\hat{Z}_i^k = 1} e^{-\overline{Z}_i^k})\\
    \mathcal{L} =& \sum_i \mathcal{L}_i
\end{aligned}
\end{equation}
where $\overline{Z}_i$ is a flattened version of $Z_i$, and $\hat{Z}_i$ denotes the flattened ground truth, containing only 1 and 0.

\subsubsection{Explicit Schema Instructor}
\label{subsec:query}

The $i$-th query $Q_i$ consists of an Explicit Schema Instructor (ESI) and the text $\mathbf{x}$. ESI is a concatenation of a prefix $p_i$ and types $t_i = [t_i^1, t_i^2, \dots]$.
The prefix $p_i$ models $(\mathbf{s, t})_{<i}$ in Equation \ref{new UIE}, which is constructed based on the sequence of previously extracted types and the corresponding sequence of spans. $t_i$ specifies what types can be potentially identified from $\mathbf{x}$ given $p_i$.

We insert a special token \texttt{[P]} before each prefix and a \texttt{[T]} before each type. Additionally, we insert a token \texttt{[Text]} before the text $\mathbf{x}$. Then, the input $Q_i$ can be represented as
\begin{equation}
    Q_i = \texttt{[CLS]} \texttt{[P]} p_i \texttt{[T]} t_i^1 \texttt{[T]} t_i^2 \dots \texttt{[Text]} x_0 x_1 \dots
\end{equation}


The biggest difference between ESI and implicit schema instructor is that the sub-types that each type can undertake are explicitly specified. Given the parent type, the semantic meaning of each sub-type is richer, thus the RexUniNLU has a better understanding to the labels.

Some detailed examples of ESI are listed in Appendix \ref{sec:show case}.

\subsubsection{Token Linking Operations}
\label{subsec:token linking}

Given the calculated score matrix $Z$, we obtain $\tilde{Z}$ from $Z$ by a predefined threshold $\delta$ following
\begin{equation}
    \begin{aligned}
        \tilde{Z}^{i,j} = \left \{ \begin{matrix}
        1 & \text{if $Z^{i,j} \ge \delta $}\\
        0 & \text{otherwise}
        \end{matrix} \right .
    \end{aligned}
\end{equation}

Token linking is performed on $\tilde{Z}$ , which takes binary values of either 1 or 0 \cite{usm}. A token linking is established from the 
$i$-th token to the $j$-th token only if $\tilde{Z}^{i,j} = 1; $ otherwise, no link exists. To illustrate this process, consider the example depicted in Figure \ref{fig: sample}. We expound upon how entities and relations can be extracted based on the score matrix.

\paragraph{Token Head-Tail Linking} Token head-tail linking serves the purpose of span detection. if $i \le j$ and $\tilde{Z}^{i,j} = 1$, the span $Q^{i:j}$ should be extracted.
The orange section in Figure \ref{fig: sample} performs token head-tail linking, wherein both ``Steve Jobs'' and ``Apple'' are recognized as entities. Consequently, a connection exists from ``Steve'' to ``Jobs'' and another from ``Apple'' to itself.

\paragraph{Token Head-Type Linking} Token head-type linking refers to the linking established between the head of a span and its type. To signify the type, we utilize the special token \texttt{[T]}, which is positioned just before the type token. As highlighted in the green section of Figure \ref{fig: sample}, ``Steve Jobs'' qualifies as a ``person'' type span, so a link points from ``Steve'' to the \texttt{[T]} token that precedes ``person''. Similarly, a link exists from ``Apple'' to the \texttt{[T]} token preceding ``org''.

\paragraph{Type-Token Tail Linking} Type-token tail linking refers to the connection established between the type of a span and its tail. Similar to token head-type linking, we utilize the \texttt{[T]} token before the type token to represent the type.
As highlighted in the blue section of Figure \ref{fig: sample}, a link exists from the \texttt{[T]} token preceding ``person'' to ``Jobs'' due to the prediction that ``Steve Jobs'' is a ``person'' span.

During inference, for a pair of token $\left \langle i, j \right \rangle$, if $Z^{i,j} \ge \delta$, and there exists a \texttt{[T]} $k$ that satisfies $Z^{i,k} \ge \delta$ and $Z^{k,j} \ge \delta$, we extract the span $Q^{i:j}$ with the type after $k$.

\subsubsection{Prompts Isolation}

\begin{figure}
    \centering
    \includegraphics[width=0.7\linewidth]{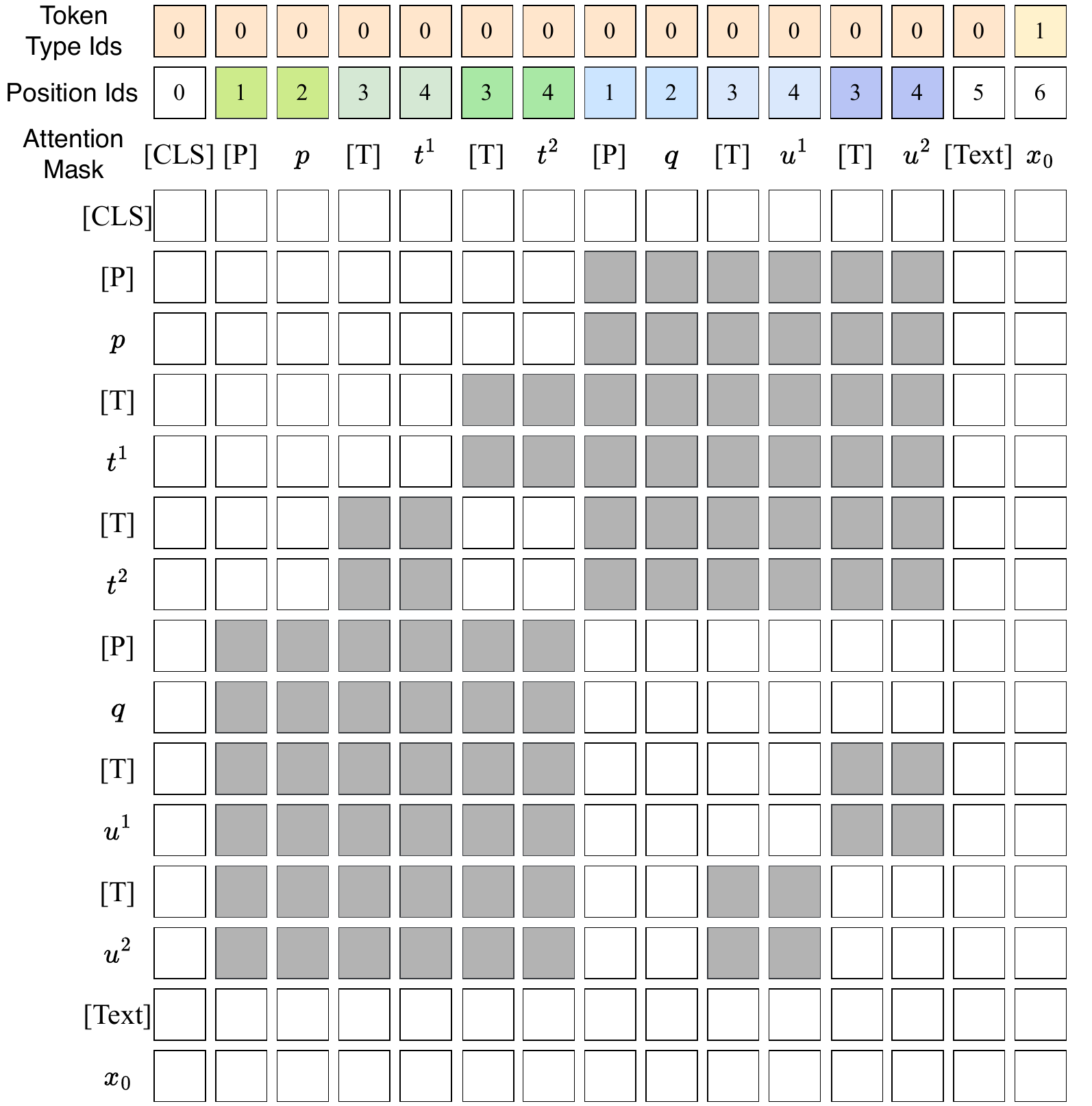}
    \caption{Token type ids, position ids and Attention mask for RexUniNLU. $p$ and $t$ denote the prefix and types of the first group of previously extracted results. $q$ and $u$ denote the prefix and types for the second group.}
    \label{fig:unimc}
\end{figure}

RexUniNLU can receives queries with multiple prefixes. To save the cost of time, we put different prefix groups in the same query.
For instance, consider the text ``Kennedy was fatally shot by Lee Harvey Oswald on November 22, 1963'', which contains two ``person'' entities. We concatenate the two entity spans, along with their corresponding types in the schema respectively to obtain ESI: \textit{[CLS][P]person: Kennedy [T] kill (person) [T] live in (location)}$\dots$\textit{[P] person: Lee Harvey Oswald [T] kill(person) [T] live in (location)}$\dots$.

However, the hidden representations of type \textit{kill (person)} should not be interfered by type \textit{live in (location)}. Similarly, the hidden representations of prefix \textit{person: Kennedy} should not be interfered by other prefixes (such as \textit{person: Lee Harvey Oswald}) either.

Inspired by UniMC \cite{unimc}, we present Prompts Isolation, an approach that mitigates interferences among tokens of diverse types and prefixes. By modifying token type ids, position ids, and attention masks, the direct flow of information between these tokens is effectively blocked, enabling clear differentiation among distinct sections in ESI.
We illustrate Prompts Isolation in Figure \ref{fig:unimc}. For the attention masks, each prefix token can only interact with the prefix itself, its sub-type tokens, and the text tokens. Each type token can only interact with the type itself, its corresponding prefix tokens, and the text tokens.

Then the position ids $P$ and attention mask $M$ in Equation \ref{equation: rotary} can be updated.
In this way, potentially confusing information flow is blocked. Additionally, the model would not be interfered by the order of prefixes and types either.

\begin{table}
    \centering
    \footnotesize
    \caption{pre-training data distribution}
    \label{tab:dist}
    \begin{tabular}{ccc}
        \toprule
         & \textbf{Task Type} & \textbf{Sample Num}\\
         \midrule
        $\mathcal{D}_{distant}$ & NER,RE & 9,619,421 \\
         \midrule
         \multirow{9}{*}{$\mathcal{D}_{superv}$} & NER & 3,893,190\\
         & RE & 1,038,111\\
         & EE & 338,215\\
         & ABSA & 2,053,980\\
         & CLS & 5,826,573\\
         & Sentiment & 289,249\\
         & Match & 1,787,641\\
         & NLI & 1,468,865\\
         & MRC & 1,449,080\\
         \midrule
         Total &  & 27,764,325 \\
         \bottomrule
    \end{tabular}
\end{table}

\begin{table}[]
    \centering
    \caption{Coverage of the tasks, datasets and settings (``Full Shot'', ``Few Shot'' and ``Zero Shot'') in this paper.}
    \label{tab:task cover}
    \begin{tabular}{cccccc}
    \toprule
    & \textbf{Task} & \textbf{Dataset} & \textbf{Full} & \textbf{Few} & \textbf{Zero} \\
    \midrule
        \multirow{7}{*}{IE} & NER & \cite{cblue,jie2019better} & \checkmark & \checkmark & \checkmark \\
        & RE & \cite{ace2005,conll04,nyt,scierc,coae2016} & \checkmark & \checkmark & \checkmark \\
        & EE & \cite{ace2005,casie,ccks} & \checkmark & \checkmark & \checkmark \\
        & MRC & \cite{pclue,cmrc2018} & \checkmark & \checkmark & \checkmark \\
        & ABSA & \cite{14res,15res,16res} & \checkmark & \checkmark & \checkmark \\
        & Quadruple & \cite{chia-etal-2022-dataset} & \checkmark & & \\
        & Quintuple & \cite{liu-etal-2021-comparative} & \checkmark & & \\
        \midrule
        \multirow{5}{*}{\shortstack{General\\CLS}} & CLS & \cite{pclue,toutiao} & \checkmark & \checkmark & \checkmark \\
         & Sentiment & \cite{nlpcc14} & \checkmark & \checkmark & \checkmark \\
         & Match & \cite{afqmc} & \checkmark & \checkmark & \checkmark \\
         & NLI & \cite{pclue,ocnli} & \checkmark & \checkmark & \checkmark \\
         & MRC & \cite{c3} & \checkmark & \checkmark & \checkmark \\
    \bottomrule
    \end{tabular}
    
\end{table}

\begin{table}[th]
    \centering
    \caption{Models compared in the experiments.}
    \label{tab:baseline}
    \begin{tabular}{cccc}
    \toprule
    \textbf{Model} & \textbf{IE} & \textbf{NLU} & \textbf{PTM} \\
    \midrule
        T5-UIE \cite{duuie} & \checkmark && T5-Large \\
        USM \cite{usm} & \checkmark && RoBERTa-Large \\
        UniMC \cite{unimc} && \checkmark & RoBERTa-wwm-ext \\
        UTC \cite{utc} &&\checkmark & ERNIE-Base\\
        PromptCLUE \cite{promptclue} & \checkmark & \checkmark & mT5-Base\\
        mT5-ZSAC \cite{zsac} & \checkmark & \checkmark & mT5-Base\\
        SiameseUniNLU \cite{siameseuninlu} & \checkmark & \checkmark & RoBERTa-wwm-ext\\
        RexUniNLU & \checkmark & \checkmark & Deberta-v2-Base/Large\\
    \bottomrule
    \end{tabular}
\end{table}

\begin{table*}[th]
    \centering
    \caption{Full-Shot results of RexUniNLU.}
    \label{tab:fullshot nlu}
    \begin{tabular}{cc|ccccc|c|c}
    \toprule
    \textbf{Dataset} & \textbf{Task} & \textbf{UniMC} & \textbf{UTC} & \textbf{PromptCLUE} & \textbf{mT5-ZSAC} & \textbf{SiameseUniNLU} & \textbf{RexUniNLU$_{\text{Base}}$} & \textbf{RexUniNLU$_{\text{Large}}$} \\
    \midrule
        CMeEE-NER \cite{cblue} & NER & - & - & 46.47 & 52.54 & 65.41 & \underline{68.39} & \textbf{70.66}\\
        Youku \cite{jie2019better} & NER & - & - & 68.35 & 71.77 & 87.16 & \underline{87.84} & \textbf{88.62}\\
        CoAE2016 \cite{coae2016} & RE & - & - & 50.68 & 60.62 & \underline{65.66} & 65.63 & \textbf{73.74}\\
        CCKS-EE \cite{ccks} & EE & - & - & 8.92 & 10.25 & 37.50 & \textbf{62.00} & \underline{55.17}\\
        CMRC2018 \cite{cmrc2018} & MRC & - & - & 53.15 & \textbf{59.43} & 46.36 & \underline{59.35} & 58.02\\
        TouTiao \cite{toutiao} & CLS & 84.11 & 91.63 &  91.64 & \underline{91.73} & 91.02 & \textbf{92.73} & 91.69\\
        NLPCC14-SC \cite{nlpcc14} & Sentiment & 83.90 & 85.50 &  81.90 & 82.50 & 85.15 & \textbf{85.90} & \underline{85.80}\\
        OCNLI \cite{ocnli} & NLI & 35.76 & \textbf{78.85} & 71.25 & 73.86 & 73.77 & 77.29 & \underline{78.47} \\
        AFQMC \cite{afqmc} & Match & 45.53 & 75.65 & 71.87 & 73.54 & 74.25 & \textbf{80.32} & \underline{76.73}\\
        $\text{C}^3$ \cite{c3} & MRC & 64.05 & 67.16 & 64.47 & 62.26 & 55.87 & \underline{68.62} & \textbf{75.56}\\
    \midrule
        \multirow{3}{*}{Average} & IE & - & - & 45.51 & 50.92 & 60.42 & \underline{68.64} & \textbf{69.24}\\
        & CLS & 62.67 & 74.36 & 76.23 & 76.78 & 76.01 & \underline{80.97} & \textbf{81.65}\\
        & All & - & - & 60.87 & 63.85 & 68.22 & \underline{74.81} & \textbf{75.45}\\
    \bottomrule
    \end{tabular}
\end{table*}

\begin{table*}[]
    \centering
    \caption{Low-resource results of RexUniNLU.}
    \label{tab:fewshot nlu}
    \begin{tabular}{cc|ccccc|c|c}
    \toprule
    \textbf{Dataset} & \textbf{Shot} & \textbf{UniMC} & \textbf{UTC} & \textbf{PromptCLUE} & \textbf{mT5-ZSAC} & \textbf{SiameseUniNLU} & \textbf{RexUniNLU$_{\text{Base}}$} & \textbf{RexUniNLU$_{\text{Large}}$} \\
    \midrule
        PCLUE-CLS-PUB & 0 &-&-&51.57 & \textbf{69.62} & 41.89 & \underline{66.50} & 65.96 \\
        PCLUE-NLI-PUB & 0 &-&-&37.29 &62.16 & \underline{70.64} & 70.62 & \textbf{76.88} \\
        PCLUE-MRC-PUB & 0 &-&-&43.63 &39.02 & 47.98 & \underline{74.39} & \textbf{77.13} \\
        \midrule
        \multirow{4}{*}{CMeEE-NER} & 0 & - & -& 0 & 0 & 52.61 & \textbf{60.62} & \underline{58.51}\\
         & 1 & - &- & 2.47 & 3.88 & 53.48 & \textbf{60.62} & \underline{59.16}\\
         & 5 & - & -& 13.66 & 12.74 & 54.15 & \underline{61.60} & \textbf{64.59}\\
         & 10 & - &- & 19.56 & 22.84 & 62.42 & \underline{63.29} & \textbf{66.47}\\
        \midrule
        \multirow{4}{*}{Youku} & 0 & - & -& 25.64 & 31.49 & \underline{49.90} & 49.64 & \textbf{53.07}\\
         & 1 & - &- & 28.81 & 30.92 & 55.19 & \underline{55.98} & \textbf{58.27}\\
         & 5 & - & -& 33.53 & 36.75 & 51.64 & \underline{59.30} & \textbf{65.63}\\
         & 10 & - &- & 44.38 & 51.70 & 62.84 & \underline{66.93} & \textbf{71.82}\\
        \midrule
        \multirow{4}{*}{CoAE2016} & 0 & - & -& 0 & 0 & 29.80 & \textbf{44.93} & \underline{43.61}\\
         & 1 & - &- & 0 & 0 & 35.90 & \underline{46.02} & \textbf{52.63}\\
         & 5 & - & -& 0 & 25.74 & \underline{50.00} & 49.28 & \textbf{55.61}\\
         & 10 & - &- & 14.82 & 37.61 & \underline{54.73} & 52.46 & \textbf{68.06}\\
         \midrule
          \multirow{4}{*}{CCKS-EE} & 0 & - & -& 0 & 0 & 2.63 & \underline{13.51} & \textbf{17.28}\\
         & 1 & - &- & 0 & 2.08 & 9.71 & \underline{30.61} & \textbf{33.04}\\
         & 5 & - & -& 4.46 & 8.75 & 43.96 & \underline{48.94} & \textbf{53.91}\\
         & 10 & - &- & 8.18 & 11.36 & 36.56 & \textbf{58.49} & \underline{56.91}\\
         \midrule
         \multirow{4}{*}{CMRC2018} & 0 & - &- & 51.17 & 50.24 & 41.81 & \underline{60.55} & \textbf{61.83}\\
         & 1 & - &- & 52.42 & 57.13 & 41.98 & \textbf{63.56} & \underline{63.42}\\
         & 5 & - &- & 50.66 & 56.48 & 41.91 & \underline{60.59} & \textbf{64.74}\\
         & 10 & - &- & 50.81 & 54.94 & 43.02 & \underline{61.67} & \textbf{65.63}\\
         \midrule
         \multirow{4}{*}{TouTiao} & 0 & 83.51 & 76.14 & 67.30 & 75.95 & 80.10 & \underline{83.87} & \textbf{85.05}\\
         & 1 & 80.80 & 78.00 & 73.08&\underline{85.00} &81.43 &84.94 &\textbf{86.03}\\
         & 5 & 84.20 &78.78&77.03&84.15&81.78&\textbf{85.65}&\underline{85.31}\\
         & 10 & \textbf{88.46} &80.73&79.52&86.45&83.16&\underline{86.94}&86.39\\
         \midrule
         \multirow{4}{*}{NLPCC14-SC} & 0 &70.50&60.10&62.25&77.90&82.36&\textbf{83.70}&\underline{83.21}\\
         & 1 &68.80&79.20&63.79&77.90&82.36&\textbf{83.70}&\underline{83.59}\\
         & 5 &\textbf{86.80}&80.60&65.39&78.80&82.36&\underline{83.70}&83.21\\
         & 10 &\textbf{90.60}&80.50&66.25&79.80&82.36&\underline{83.70}&83.21\\
         \midrule
         \multirow{4}{*}{OCNLI} & 0 &31.22&41.36&30.71&31.32&69.31&\underline{75.73}&\textbf{79.73}\\
         & 1 &51.05&71.90&32.30&33.29&69.64&\underline{76.57}&\textbf{80.51}\\
         & 5 &70.17&71.29&34.47&37.86&69.31&\underline{77.29}&\textbf{80.85}\\
         & 10 &70.47&70.98&36.13&38.51&69.69&\underline{77.40}&\textbf{80.71}\\
         \midrule
        \multirow{4}{*}{AFQMC} & 0 &30.93&24.56&43.68&69.00&50.59&\textbf{79.88}&\underline{76.52}\\
         & 1 &54.96&75.09&44.16&69.00&67.36&\textbf{79.88}&\underline{79.43}\\
         & 5 &74.56&75.28&46.82&69.09&70.92&\textbf{82.36}&\underline{79.43}\\
         & 10 &68.10&74.44&50.05&69.02&67.41&\textbf{82.25}&\underline{79.05}\\
         \midrule
         \multirow{4}{*}{$\text{C}^3$} & 0 &55.74&60.27&52.34&51.45&31.56&\underline{67.01}&\textbf{74.83}\\
         & 1 & 61.89&65.75&54.33&52.12&36.53&\underline{67.13}&\textbf{74.83}\\
         & 5 & 59.13&66.19&54.53&51.66&36.60&\underline{67.01}&\textbf{74.83}\\
         & 10 & 62.06&66.17&55.38&51.81&36.15&\underline{67.01}&\textbf{74.83}\\
         \midrule
         \multirow{4}{*}{Average} & 0 &-&-&33.35&38.74&49.07&\underline{61.95}&\textbf{63.37}\\
         & 1 &-&-&35.14&41.13&53.36&\underline{64.90}&\textbf{67.09}\\
         & 5 &-&-&38.06&46.20&58.26&\underline{67.57}&\textbf{70.82}\\
         & 10 &-&-&42.51&50.41&59.83&\underline{70.02}&\textbf{73.31}\\
    \bottomrule
    \end{tabular}
\end{table*}

\subsection{RexUniNLU for CLS}\label{sec:CLS}


Although SiameseUniNLU \cite{wang2019structbert,Zhao2021AdjacencyLO} can process text classification and IE tasks, there are three key weaknesses for classification: 1) There is a risk that the extraction framework predicts no classes if the threshold is set too high. 2) As labels share the length limitation with the input sentence, there may be information to be truncated when the labels are too long. 3) SiameseUniNLU cannot classify sentences with hierarchical labels.

We insert two kinds of special tokens ahead the input sequence, \texttt{[CLASSIFY]} and \texttt{[MULTICLASSIFY]}, to indicate classification and multi-label classification tasks, and represent the whole input sequence, which play the role of extraction span to build connections between the input sequence and the label text. The score matrix is illustrated in Figure \ref{fig:cls-metrix}, where \texttt{[CLST]} could be either \texttt{[CLASSIFY]} or \texttt{[MULTICLASSIFY]}.

The span to be extracted is fixed to the special token \texttt{[CLST]}. However, different from IE, classification requires at least one extracted label. So we have proposed a hand-shaking mechanism.

The score matrix is activated to $[0, 1]$:
\begin{equation}
    \widehat{Z}_i = \mathbf{Sigmoid}(Z_i)
\end{equation}

For single-label classification, we calculate the label with maximum pair logits. Let $j$ denote the position of \texttt{[CLASSIFY]}:
\begin{equation}
    y = \arg\max_{y}(\widehat{Z}_i^{j,y} \times \widehat{Z}_i^{y,j})
\end{equation}

For multi-label classification, we add a constraint that the score of each cell should be larger than a threshold $\delta$ (set as 0.9 for implementation). Let $j$ denote the position of \texttt{[MULTICLASSIFY]}:
\begin{equation}
    y = \{y' | \widehat{Z}_i^{j,y'} > \delta, \widehat{Z}_i^{y',j}>\delta\}
\end{equation}

If there are too many labels or the label texts are too long, we fix the maximum length and construct iterative queries with the truncated text segments, and obtain the final results through ensemble methods. The leverage of recursive queries also handles the hierarchical classification tasks.

\section{Experiments}

\subsection{Pre-Training}

To enhance the zero-shot and few-shot performance of RexUniNLU, we pre-trained RexUniNLU on the following two distinct datasets:

\paragraph{Distant Supervision data $\mathcal{D}_{distant}$} We gathered the corpus and labels from Chinese WikiPedia\footnote{https://zh.wikipedia.org/} and Baidu Encyclopedia\footnote{https://baike.baidu.com/}, and utilized Distant Supervision to align the texts with their respective labels. We remove abstract and over-specialized entity types and relation types (such as ``structural class of chemical compound'') and remove categories that occur less than 10000 times. We also remove the examples that do not contain any relations. Finally, we collect over 9M samples containing entities and relations as $\mathcal{D}_{distant}$.

\paragraph{Supervised NLU data $\mathcal{D}_{superv}$} Compared with $\mathcal{D}_{distant}$, supervised data exhibits higher quality due to its absence of abstract or over-specialized classes, and there is no high false negative rate caused by incomplete knowledge base. We collect over 18M high-quality supervised Chinese data for all NLU tasks from NLPCC\footnote{http://tcci.ccf.org.cn/nlpcc.php}, CAIL\footnote{http://cail.cipsc.org.cn/}, CCKS\footnote{https://www.cipsc.org.cn/}, THUCNews\footnote{http://thuctc.thunlp.org/}, MSRA\cite{levow-2006-third}, etc. The detail of our pre-training data distribution is shown in Table \ref{tab:dist}.

\paragraph{English data $\mathcal{D}_{en}$} As mentioned, RexUniNLU was supposed to generalize to all IE tasks including those with quadruples and quintuples, which aim at extracting schema with three or more spans. However, no high-quality Chinese open-source Quadruple Extraction and Quintuple Extraction dataset were found. Besides, we also want to validate the generalization capability of RexUniNLU under different languages. Therefore, we pretrained an English version RexUniNLU on IE tasks, namely RexUIE-EN. 

Similarly, we gathered 3M distant supervision data from WikiPedia\footnote{https://www.wikipedia.org/} with the same pre-processing method. Also, we employ OntoNotes \cite{pradhan-etal-2013-towards}, NYT \cite{nyt}, CrossNER \cite{liu2020crossner}, Few-NERD \cite{ding-etal-2021-nerd}, kbp37 \cite{DBLP:journals/corr/ZhangW15a}, Mit Restaurant and Movie corpus \cite{mitrest} together as supervised data. Besides, since the empirical results of previous works \cite{usm} show that incorporating machine reading comprehension (MRC) data into pre-training enhances the model's capacity to utilize semantic information in prompt. Accordingly, we collect SQuAD \cite{2016arXiv160605250R} and HellaSwag \cite{zellers2019hellaswag} together as MRC supervised instances to the pre-training stage. The MRC data is constructed with pairs of questions and answers, and we use the questions as the type, and answers as the span to extract for implementations. 

\subsection{Test Dataset}

The test tasks and datasets covered in this paper are listed in Table \ref{tab:task cover}. For English IE, we mainly follow the data setting of previous works \cite{duuie, usm}. We add two more English tasks to evaluate the ability of extracting schemas with more than two spans: HyperRED \cite{chia-etal-2022-dataset} and Comparative Opinion Quintuple Extraction \cite{liu-etal-2021-comparative} (COQE).

\subsection{Models Comparison}


We adopt models designed for IE-only tasks, NLU-only tasks and the models for both IE and NLU tasks. The models are listed in Table \ref{tab:baseline}.
The detailed introduction to the models are available in Section \ref{sec:related work}.
Due to fact that the baselines are at different scales, we conduct the pre-training to produce RexUniNLU at two scales, thus there are Base version and Large version, respectively.

\subsection{Results of RexUniNLU}

\subsubsection{Full-shot Results}

We first undertook experiments with full-shot training data foundation on pre-trained models across all Chinese NLU tasks. Table \ref{tab:fullshot nlu} presents a comprehensive comparison of RexUniNLU$_{\text{Base}}$ and RexUniNLU$_{\text{Large}}$ against UniMC \cite{unimc}, UTC \cite{utc}, PromptCLUE \cite{promptclue}, mT5-ZSAC \cite{zsac} and SiameseUniNLU \cite{siameseuninlu}. 
The results exhibits that: 1) For models with comparable scale, RexUniNLU$_{\text{Base}}$ obtained the best performance over IE and CLS tasks. Furthermore, the performance of RexUniNLU$_{\text{Base}}$ shows great stability, which exhibited best or second-best score on almost all the testing data. 2) If the RexUniNLU model scaled up, the performance can be further improved, especially for those hard tasks such as CMeEE-NER, CoAE2016, and $\text{C}^3$. It is worth noting that the decrease between RexUniNLU$_{\text{Base}}$ and RexUniNLU$_{\text{Large}}$ under CCKS-EE is probably because of overfitting caused by small training data scale. 3) The decoder-based generation models such as PromptCLUE and mT5-ZSAC showed poor performance on IE tasks as a result of hallucination and limited instruction following capability, which is aligned with our demonstration forward.

\subsubsection{Low-Resource Results}
Apart from full-shot settings, we conducted zero-shot and few-shot learning experiments on all Chinese NLU tasks, as shown in Table \ref{tab:fewshot nlu}. We defined the n-shot setting as each label has to be presented in the training set for n times \cite{duuie}. According to the results, we found out that 1) RexUniNLU$_{\text{Base}}$ and RexUniNLU$_{\text{Large}}$ apparently outperform all the baseline models, they obained best or second best score on almost all the test data and settings. 2) In comparison to mT5-ZSAC \cite{zsac} and SiameseUniNLU \cite{siameseuninlu}, which possess similar model sizes and were pre-trained on data scales of 30M and 27M respectively, RexUniNLU$_{\text{Base}}$ demonstrates superior performance in zero-shot settings. Over 26.25\% performance gain was obtained from SiameseUniNLU to RexUniNLU$_{\text{Base}}$, indicateing a much higher generalization capability of Rex-framework. 3) RexUniNLU can adapt to low-resource scenarios well. For instance, given the strong zero-shot capability, the performance of RexUniNLU$_{\text{Base}}$ can be continuingly improved by 4.76\%, 9.07\% and 13.02\% when the 1-shot, 5-shot and 10-shot training data were provided. What's more, the law-resource performance can be further improved when change RexUniNLU$_{\text{Base}}$ to RexUniNLU$_{\text{Large}}$. 4) The advantage of RexUniNLU was more significant for those harder tasks like IE and MRC. When compared with the suboptimal model SiameseUniNLU, RexUniNLU$_{\text{Base}}$ was 42.22\%, 39.16\%, 24.60\%, 25.08\% higher in average for 0-shot, 1-shot, 5-shot and 10-shot settings across all IE and MRC tasks. Even for those simpler CLS tasks, RexUniNLU$_{\text{Base}}$ was also 59.85\%, 6.86\%, 7.82\%, 8.02\% higher in average for 0-shot, 1-shot, 5-shot and 10-shot settings over current SOTA unified text classification model UTC \cite{utc}.

\subsection{Results of RexUIE-EN}

\subsubsection{Full-shot Results}
\begin{table}[t]
    \centering
    \tiny
    \caption{F1 result for UIE models with pre-training. $*\text{-Trg}$ means evaluating models with Event Trigger F1, $*\text{-Arg}$ means evaluating models with Event Argument F1. T5-UIE \cite{duuie} and USM \cite{usm} are the previous SoTA UIE models. ``TSS'' denotes Task-specific SoTAs.}
    \label{tab:main_result}
    \begin{tabular}{c|c|m{0.6cm}<{\centering}m{0.2cm}<{\centering}m{1cm}<{\centering}|m{0.6cm}<{\centering}m{0.2cm}<{\centering}m{1cm}<{\centering}}
    \toprule
        \multirow{2}*{\textbf{Dataset}} & \multirow{2}*{\textbf{TSS}} & \multicolumn{3}{c|}{Without Pre-training} & \multicolumn{3}{c}{With Pre-training}\\
        & & \textbf{T5-UIE} & \textbf{USM} & \textbf{RexUIE-EN} & \textbf{T5-UIE} & \textbf{USM} & \textbf{RexUIE-EN} \\
    \midrule
        ACE04 & \textbf{87.90} & 86.52 & 87.79 & \textbf{88.02} & 86.89 & 87.62 & 87.25\\
        ACE05-Ent & 86.91 & 85.52 & \textbf{86.98} & 86.87 & 85.78 & 87.14 & \textbf{87.23}\\
        CoNLL03 & 93.21 & 92.17 & 92.79 & \textbf{93.31} & 92.99 & 93.16 & \textbf{93.67}\\
        \midrule
        ACE05-Rel & \textbf{66.80} & 64.68 & 66.54 & 63.44 & 66.06 & \textbf{67.88} & 64.87\\
        CoNLL04 & 75.40 & 73.07 & 75.86 & \textbf{76.79} & 75.00 & \textbf{78.84} & 78.39\\
        NYT & 93.40 & 93.54 & 93.96 & \textbf{94.35} & 93.54 & 94.07 & \textbf{94.55} \\
        SciERC & \textbf{38.40} & 33.36 & 37.05 & 38.16 & 36.53 & 37.36 & 38.37\\
        \midrule
        \shortstack{ACE05-Evt-Trg} & \textbf{73.60} & 72.63 & 71.68 & 73.25 & 73.36 & 72.41 & \textbf{75.17}\\
        \shortstack{ACE05-Evt-Arg} &55.10 & 54.67 & 55.37 & \textbf{57.27} & 54.79 & 55.83 & \textbf{59.15}\\
        \shortstack{CASIE-Trg} & 68.98 & 68.98 & 70.77 & \textbf{72.03} & 68.33 & 71.73 & \textbf{73.01}\\
        \shortstack{CASIE-Arg} & 60.37 & 60.37 & \textbf{63.05} & 62.15 & 61.30 & 63.26 & \textbf{63.87}\\
        \midrule
        14-res & 72.16 & 73.78 & 76.35 & \textbf{76.36} & 74.52 & 77.26 & \textbf{77.46}\\
        14-lap &60.78 & 63.15 & 65.46 & \textbf{66.92} & 63.88 & 65.51 & \textbf{66.41}\\
        15-res & 63.27 & 66.10 & 68.80 & \textbf{70.48} & 67.15 & 69.86 & \textbf{70.84}\\
        16-res & 70.26 & 73.87 & 76.73 & \textbf{78.13} & 75.07 & \textbf{78.25} & 77.20\\
        \midrule
        HyperRED & 66.75 & - & - & \textbf{73.25} & - & - & \textbf{75.20} \\
        \midrule
        Camera-COQE & 13.36 & - & - & \textbf{32.02} & - & - & \textbf{32.80}\\
    \bottomrule
    \end{tabular}
\end{table}

We then conduct experiments with full-shot training data over English IE tasks. Table \ref{tab:main_result} presents a comprehensive comparison of RexUIE-EN against T5-UIE \cite{duuie}, USM \cite{usm}, and previous task-specific models, both in pre-training and non-pre-training scenarios.
We can observe that: 1) RexUIE-EN surpasses the task-specific state-of-the-art models on more than half of the IE tasks even without pre-training. RexUIE-EN exhibits a higher F1 score than both USM and T5-UIE across all the ABSA datasets. Furthermore, RexUIE-EN's performance in the task of Event Extraction is remarkably superior to that of the baseline models.
2) Pre-training brings in slight performance improvements. By comparing the outcomes in the last three columns, we can observe that RexUIE-EN with pre-training is ahead of T5-UIE and USM on the majority of datasets.
After pre-training, ACE05-Evt showed a significant improvement with an approximately 2\% increase in F1 score. This implies that RexUIE-EN effectively utilizes the semantic information in prompt texts and establishes links between text spans and their corresponding types. It is worth noting that the schema of trigger words and arguments in ACE05-Evt is complex, and the model heavily relies on the semantic information of labels.
3) The bottom two rows describe the results of extracting quadruples and quintuples, and they are compared with the SoTA methods. Our model demonstrates significantly superior performance on both HyperRED and Camera-COQE, which shows the effectiveness of extracting complex schemas.

\subsubsection{Low-Resource Results}

\begin{table}[t]
    \centering
    \caption{Few-Shot experimental results. AVE-S denotes the average performance over 1-Shot, 5-Shot and 10-Shot.}
    \label{tab:fewshot}
    \begin{tabular}{cccccc}
    \toprule
        & \textbf{Model} & \textbf{1-Shot} & \textbf{5-Shot} & \textbf{10-Shot} & \textbf{AVE-S}\\
        \midrule
         \multirow{3}{*}{\shortstack{Entity \\ CoNLL03}} & T5-UIE & 57.53 & 75.32 & 79.12 & 70.66 \\
         & USM & 71.11 & 83.25 & 84.58 & 79.65 \\
         & RexUIE-EN & \textbf{86.57} & \textbf{89.63} & \textbf{90.82} & \textbf{89.07}\\
         \midrule
        \multirow{3}{*}{\shortstack{Relation \\ CoNLL04}} & T5-UIE & 34.88 & 51.64 & 58.98 & 48.50 \\
         & USM & 36.17 & 53.2 & 60.99 & 50.12 \\
         & RexUIE-EN & \textbf{43.80} & \textbf{54.90} & \textbf{61.68} & \textbf{53.46}\\
         \midrule
        \multirow{3}{*}{\shortstack{Event \\ Trigger \\ ACE05-Evt}} & T5-UIE & 42.37 & 53.07 & 54.35 & 49.93 \\
         & USM & 40.86 & 55.61 & 58.79 & 51.75 \\
         & RexUIE-EN & \textbf{56.95} & \textbf{64.12} & \textbf{65.41} & \textbf{62.16}\\
         \midrule
        \multirow{3}{*}{\shortstack{Event \\ Argument \\ ACE05-Evt}} & T5-UIE & 14.56 & 31.20 & 35.19 & 26.98 \\
         & USM & 19.01 & 36.69 & 42.48 & 32.73 \\
         & RexUIE-EN & \textbf{30.43} & \textbf{41.04} & \textbf{45.14} & \textbf{38.87}\\
         \midrule
        \multirow{3}{*}{\shortstack{Sentiment \\ 16-res}} & T5-UIE & 23.04 & 42.67 & 53.28 & 39.66 \\
         & USM & 30.81 & \textbf{52.06} & 58.29 & 47.05 \\
         & RexUIE-EN & \textbf{37.70} & 49.84 & \textbf{60.56} & \textbf{49.37}\\
        \bottomrule
    \end{tabular}
\end{table}

We conducted few-shot experiments on one dataset for each task, following \cite{duuie} and \cite{usm}.
The results are shown in Table \ref{tab:fewshot}.
In general, RexUIE-EN exhibits superior performance compared to T5-UIE and USM in a low-resource setting.  Specifically, RexUIE-EN relatively outperforms T5-UIE by 56.62\% and USM by 32.93\% on average in 1-shot scenarios. 
The success of RexUIE-EN in low-resource settings can be attributed to its ability to extract information learned during pre-training, and to the efficacy of our proposed query, which facilitates explicit schema learning by RexUIE-EN.
\begin{table}[t]
    \centering
    \caption{Zero-Shot performance on RE and NER. * indicates that the experiment is conducted by ourselves.}
    \label{tab:zero-shot}
    \begin{tabular}{cccccc}
    \toprule
        \textbf{Task} & \textbf{Model} & \textbf{Precision} & \textbf{Recall} & \textbf{F1}\\
    \midrule
    \multirow{5}*{\shortstack{Relation \\ CoNLL04}} & GPT-3 & - & - & 18.10 \\
       & ChatGPT* & 25.16 & 19.16 & 21.76 \\
       & D{\scriptsize EEP}S{\scriptsize TRUCT} & - & - & 25.80 \\
       & USM & - & - & 25.95 \\
       & RexUIE-EN* & 44.95 & 23.22 & \textbf{30.62}\\
    \midrule
    \multirow{2}*{\shortstack{Entity \\ CoNLLpp}} & ChatGPT &  62.30 & 55.00 & 58.40 \\
       & RexUIE-EN* & 91.57 & 66.09 & \textbf{76.77} \\
    \bottomrule
    \end{tabular}
\end{table}
\begin{table}[t]
    \centering
    \caption{Zero-shot multi-modality experiments.}
    \label{tab:mrexuninlu}
    \begin{tabular}{ccc}
    \toprule
    \textbf{Model} & \textbf{Modality} & \textbf{Entity Strict F1} \\
    \midrule
        ChatGPT & text &  48.99\\
        RexUniNLU-base & text & 34.83\\
        RexUniNLU-large & text & 40.96\\
        PPN & text+layout+image & 60.73\\
    \midrule
        \multirow{3}*{MRexUniNLU} & text & 47.02\\
        & text+layout & \underline{65.69}\\
        & text+layout+image & \textbf{66.84}\\
    \bottomrule
    \end{tabular}
\end{table}
We also conducted zero-shot experiments on RE and NER comparing RexUIE-EN with other pre-trained models, including ChatGPT. We adopt the pipeline proposed by ChatIE \cite{chatie} for ChatGPT.
We used CoNLL04 and CoNLLpp \cite{wang2019cross} for RE and NER respectively.
We report Precision, Recall and F1 in Table \ref{tab:zero-shot}.
RexUIE-EN achieves the highest zero-shot extraction performance on the two datasets.
Furthermore, we analyzed bad cases of ChatGPT.
1) ChatGPT generated words that did not exist in the original text. For example, ChatGPT output a span ``Coats Michael'', while the original text was ``Michael Coats''.
2) Errors caused by inappropriate granularity, such as ``city in Italy'' and ``Italy''.
3) Illegal extraction against the schema. ChatGPT outputs (\textit{Leningrad, located in, Kirov Ballet}), while ``Kirov Ballet'' is an organization rather than a location.

\section{Cross-Modality RexUniNLU}

\subsection{Test Dataset and Model Comparison}

As mentioned, there is few existing work on multi-modality NLU for low-resource context other than the LLMs. Therefore, we selected the same test data as the PPN model \cite{wei2023ppn}, where 20 types of documents and 2408 samples are included. We used entity strict F1 as the metrics.

Apart from PPN, we selected RexUniNLU-base, RexUniNLU-large and ChatGPT \footnote{https://openai.com/blog/chatgpt} as the baseline models as well, which was supposed to process the plain-text converted by OCR (Optical Character Recognition).

\subsection{Experiment Results}

Our training strategy empowering MRexUniNLU to flexibly handle data with various modalities. The experiment results indicates that the multi-modality information, especially layout features could be necessary for document understanding. Without supplementary information beyond plain text, even ChatGPT could not exceed the entity strict F1 score of 50. In the meantime, we have affirmed the merits of the Rex-framework. Despite the model scale of MRexUniNLU being a thousandfold smaller than ChatGPT, its performance remarkably approaches that of ChatGPT in identical plain-text settings, showcasing a truly impressive feat. Once the layout and image features included, the performance of MRexUniNLU could further be improved by 39.71\% and 42.15\%, and achieved the SOTA performance.

\section{Conclusion}

In this paper, we firstly redefine the true UIE with a formulation that covers almost all extraction and classification schemas. Then we design the framework of RexUniNLU, which recursively runs queries for all schema types.
Extensive experiments conducted on information extraction, text classification in both Chinese and English, and multi-modality, revealed the effectiveness and superiority. 

\section*{Acknowledgments}

This work was supported in part by National Key Research and Development Program of China (2022YFC3340900), National Natural Science Foundation of China (62376243, 62037001, U20A20387), the StarryNight Science Fund of Zhejiang University Shanghai Institute for Advanced Study (SN-ZJU-SIAS-0010), Alibaba Group through Alibaba Research Intern Program, Project by Shanghai AI Laboratory (P22KS00111), Program of Zhejiang Province Science and Technology (2022C01044).

\bibliographystyle{IEEEtran}
\bibliography{custom.bib}

\clearpage

{\appendices

\section{Test Dataset and Metric}

\begin{table}[h!]
    \centering
    \footnotesize
    \caption{Detailed supervised datasets and evaluation metrics for each task.}
    \label{tab:superv_data_metric}
    \begin{tabular}{ccc}
        \toprule
        \textbf{Task} & \textbf{Metric} & \textbf{Dataset}\\
        \midrule
        \multirow{5}*{NER} & Entity Strict F1 & ACE04-Ent \\
         & Entity Strict F1 & ACE05-Ent\\
         & Entity Strict F1 & CoNLL03\\
         & Entity Strict F1 & CMeEE-NER\\
         & Entity Strict F1 & Youku\\
         \midrule
         \multirow{4}{*}{RE} & Relation Strict F1 & ACE05-Rel \\
          & Relation Strict F1 & CoNLL04 \\
          & Relation Triplet F1 & NYT\\
          & Relation Strict F1 & SciERC \\
          & Relation Strict F1 & CoAE2016 \\
          \midrule
          \multirow{4}{*}{EE} & Trigger F1 & ACE05-Evt \\
           & Argument F1 & ACE05-Evt \\
          & Trigger F1 & CASIE \\
          & Argument F1 & CASIE \\
          & Strict F1 & CCKS-EE \\
          \midrule
          \multirow{3}{*}{MRC} & Strict F1 & pCLUE-MRC \\
          & Strict F1 & CMRC2018 \\
          & Strict F1 & $\text{C}^3$ \\
          \midrule
          \multirow{4}{*}{ABSA} & Sentiment Strict F1 & 14-res \\
          & Sentiment Strict F1 & 14-lap \\
          & Sentiment Strict F1 & 15-res \\
          & Sentiment Strict F1 & 16-res\\
          \midrule
          Quadruple & Quadruple Strict F1 & HyperRED \\
          \midrule
          Quintuple & Sentiment Strict F1 & Camera-COQE\\
          \midrule
          Others & Strict F1 & \\
        \bottomrule
    \end{tabular}
\end{table}

The test tasks and datasets covered in this paper are listed in Table \ref{tab:task cover}. For English IE, we mainly follow the data setting of previous works \cite{duuie, usm}. We add two more English tasks to evaluate the ability of extracting schemas with more than two spans: HyperRED \cite{chia-etal-2022-dataset} and Comparative Opinion Quintuple Extraction \cite{liu-etal-2021-comparative} (COQE).
1) Quadruple Extraction. We use HyperRED \cite{chia-etal-2022-dataset}, which is a dataset for hyper-relational extraction to extract more specific and complete facts from the text. Each quadruple of HyperRED consists of a standard relation triple and an additional qualifier field that covers various attributes such as time, quantity, and location.
2) Quintuple Extraction. Comparative Opinion Quintuple Extraction \cite{liu-etal-2021-comparative} (COQE), aims to extract all the comparative quintuples from review sentences. There are at most 5 attributes for each instance to extract: subject, object, aspect, opinion, and the polarity of the opinion(e.g. better, worse, or equal). We only use the English subset Camera-COQE.

The evaluation metrics for each dataset are listed in Table \ref{tab:superv_data_metric}. We explain the evaluation metrics as follows.
\paragraph{Entity Strict F1} An entity mention is correct if its offsets and type match a reference entity.

\paragraph{Relation Strict F1} A relation is correct if its relation type is correct and the offsets and entity types of the related entity mentions are correct.

\paragraph{Relation Triplet F1} A relation is correct if its relation type is correct and the string of the related entity mentions are correct.

\paragraph{Event Trigger F1} An event trigger is correct if its offsets and event type matches a reference trigger.

\paragraph{Event Argument F1} An event argument is correct if its offsets, role type, and event type match a reference argument mention.

\paragraph{Sentiment Strict F1} For triples, a sentiment is correct if the offsets of its target, opinion and the sentiment polarity match with the ground truth. For quintuples, a sentiment is correct if the offsets of its subject, object, aspect, opinion and the sentiment polarity match with the ground truth.

\paragraph{Quadruple Strict F1} A relation quadruple is correct if the relation type and the type and offsets of its subject, object, qualifier match with the ground truth.

\paragraph{Strict F1} The prediction is correct if it totally matches the label, which is a universal metric adaptable to all other tasks (mainly for general CLS).

\section{Pre-Training of Cross-Modality RexUniNLU}
We have diligently endeavored to gather an extensive and diverse pre-training dataset that includes annotated multi-modality information sourced from ChartQA \cite{masry2022chartqa}, DocVQA \cite{mathew2021docvqa}, KleisterCharity \cite{stanislawek2021kleister}, PWC \cite{kardas2020axcell}, WebSRC \cite{chen2021websrc}, VisualMRC \cite{tanaka2021visualmrc}, etc. Our comprehensive collection encompasses 215 distinct document types, over 416K individual documents, and in excess of 817K samples.

To achieve cross-modality processing capability, we meticulously curated a representative subset featuring 200K entries of general plain-text NLU data alongside 1M document-esque plain-text samples from the RexUniNLU training collection. We employ an alternating mix strategy for the multi-modality and plain-text training data. Specifically, we sequence every 1,200 entries of plain-text data with a subsequent of 800 multi-modality data, continuing in this pattern throughout the training set.

We developed MRexUniNLU by adopting LayoutLMv3-base \cite{huang2022layoutlmv3} as the the encoder and introducing rotary embedding \cite{su2022global} in which the Rex-framework required. We set the maximum token length of 512 and the maximum length of prompt of 256. We pre-trained MRexUniNLU with AdamW optimizer \cite{DBLP:journals/corr/abs-1711-05101}, with a weight decay ratio of 0.01, warmup ratio of 0.1 and learning rate of 1e-4.

\section{Implementation Details}

We download the supervised data for pre-training from HuggingFace\footnote{https://huggingface.co/datasets}.
For all the downstream datasets, we follow the procedure of T5-UIE and USM \cite{duuie, usm} and then convert them to the input format of RexUniNLU.
We implement the pre-training model and trainer based on Transformers \cite{wolf-etal-2020-transformers}. We adopt Erlangshen-DeBERTa-v2-97M and Erlangshen-DeBERTa-v2-310M \cite{fengshenbang} as text encoder and developed RexUniNLU-base and RexUniNLU-large respectively. For RexUIE-EN, we adopt Deberta-v3-large \cite{he2021deberta} as the text encoder. We set the maximum token length to 512, and the maximum length of prompt to 256
. We split a query into sub-queries containing prompt text segments when the length of the prompt text is beyond the limit.
Our model is optimized by AdamW \cite{DBLP:journals/corr/abs-1711-05101}, with weight decay as 0.01. The threshold $\delta$ is set as 0 for IE and 0.9 for CLS. We set the clip gradient norm as 2, warmup ratio as 0.1.



\section{Detailed Analysis}

\subsection{Complex Schema Extraction}

\begin{table}[t]
    \centering
    \caption{Extraction results on COQE.}
    \label{tab:coqe t5uie}
    \begin{tabular}{ccccc}
        \toprule
        & \textbf{Model} & \textbf{Precision} & \textbf{Recall} & \textbf{F1} \\
        \midrule
        \multirow{2}*{\scriptsize Without Pre-train} & T5-UIE & 20.92 & 21.56 & 21.23\\
        & RexUIE-EN & 31.43 & 32.62 & \textbf{32.02}\\
        \midrule
        \multirow{2}*{\scriptsize With Pre-train} & T5-UIE & 24.41 & 25.46 & 24.92\\
        & RexUIE-EN & 34.07 & 31.63 & \textbf{32.80}\\
        \bottomrule
    \end{tabular}
\end{table}

To illustrate the significance of the ability to extract complex schemas, we designed a forced approach to extract quintuples for T5-UIE, which extracts three tuples to form one quintuple.

The quintuple in COQE can be represented as (\textit{subject, object, aspect, opinion, sentiment}). We propose to model the quintuple extraction as extracting three triples for T5-UIE: (\textit{subject}, ``subject-object'', \textit{object}), (\textit{object}, ``object-aspect'', \textit{aspect}), and (\textit{aspect, sentiment, opinion}).

Table \ref{tab:coqe t5uie} shows the results comparing RexUIE-EN with T5-UIE. 
Without pre-training, the F1 scores for T5-UIE and RexUIE-EN are 21.23\% and 32.02\% respectively. After pre-training, the F1 of T5-UIE is improved to 24.92\%, while F1 of RexUIE-EN is 32.02\%.
In summary, RexUIE-EN's approach of directly extracting quintuples exhibits superior performance. Although T5-UIE shows a slight performance improvement after pre-training, it is still approximately 8\% lower than RexUIE-EN on F1.

\subsection{Absense of Explicit Schema}

\begin{figure}[thbp]
    \centering
    \includegraphics[width=\linewidth]{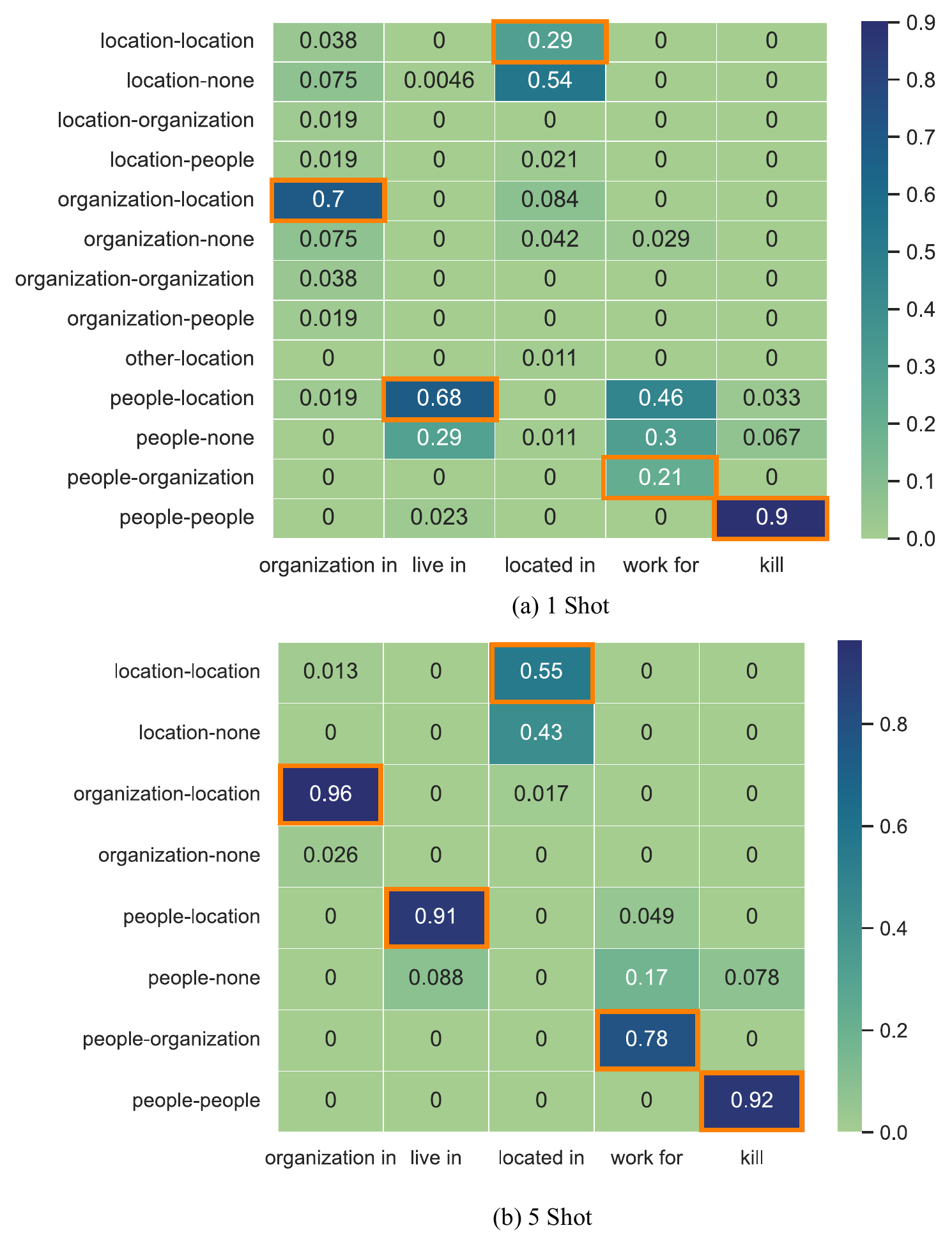}
    \caption{The distribution of relation type versus \textit{subject type}-\textit{object type} predicted by T5-UIE. We circle the correct cases in orange.}
    \label{fig:explicit schema}
\end{figure}

\begin{table}[thb]
    \centering
    \caption{Ablation Study. PI denotes Prompts Isolation, Rt denotes rotary embedding. AVG is calculated over the five tasks.}
    \label{tab:ablation study}
    \begin{tabular}{clccc}
        \toprule
         & \textbf{Method} & \textbf{Precision} & \textbf{Recall} & \textbf{F1}\\
        \midrule
        \multirow{4}*{\shortstack{Entity \\ CoNLL03}} & RexUIE-EN & 92.95 & 93.66 & 93.31 \\
         & \quad w/o PI & 93.00 & 93.79 & 93.39 \\
         & \quad w/o Rt & 92.16 & 93.84 & 92.99\\
         & \quad w/o PI+Rt & 92.37 & 93.52 & 92.94\\
         \midrule
        \multirow{4}*{\shortstack{Relation \\ CoNLL04}} & RexUIE-EN & 78.80 & 74.88 & 76.79 \\
         & \quad w/o PI & 75.31 & 72.27 & 73.76\\
         & \quad w/o Rt & 75.06 & 72.75 & 73.89\\
         & \quad w/o PI+Rt & 73.56 & 72.51 & 73.03\\
         \midrule
        \multirow{4}*{\shortstack{Event \\ Trigger \\ ACE05-Evt}} & RexUIE-EN & 71.36 & 75.24 & 73.25 \\
         & \quad w/o PI & 73.80 & 72.41 & 73.10\\
         & \quad w/o Rt & 72.06 & 76.65 & 74.29\\
         & \quad w/o PI+Rt & 73.51 & 72.64 & 73.07\\
         \midrule
         \multirow{4}*{\shortstack{Event \\ Argument \\ ACE05-Evt}} & RexUIE-EN & 56.69 & 57.86 & 57.27 \\
         & \quad w/o PI & 57.74 & 56.97 & 57.36\\
         & \quad w/o Rt & 57.93 & 59.64 & 58.77\\
         & \quad w/o PI+Rt & 55.84 & 55.34 & 55.59\\
         \midrule
        \multirow{4}*{\shortstack{Sentiment \\ 16-res}} & RexUIE-EN & 75.18 & 81.32 & 78.13 \\
         & \quad w/o PI & 78.45 & 78.60 & 78.52\\
         & \quad w/o Rt & 69.27 & 81.13 & 74.73\\
         & \quad w/o PI+Rt & 70.88 & 78.60 & 74.54\\
         \midrule
        \multirow{4}*{\shortstack{AVG}} & RexUIE-EN &  & & \textbf{75.75} \\
         & \quad w/o PI &  &  & 75.23\\
         & \quad w/o Rt &  &  & 74.93\\
         & \quad w/o PI+Rt & & & 73.83\\
         \bottomrule
    \end{tabular}
\end{table}

\begin{table}[t]
    \centering
    \caption{Replacing DeBERTa with RoBERTa. We compare $\text{RexUIE-EN}_{\text{RoBERTa}}$ with USM as they share the same encoder structure.}
    \label{tab:rexuie-roberta}
    \begin{tabular}{c|ccc}
        \toprule
        \textbf{Dataset} & \textbf{USM} & \textbf{$\text{RexUIE-EN}_{\text{RoBERTa}}$} & \textbf{RexUIE-EN}\\
        \midrule
        CoNLL03 & 92.76 & 92.98 & \textbf{93.31}\\
        CoNLL04 & 75.86 & \textbf{77.15} & 76.79\\
        {\scriptsize ACE05-Evt (Trigger)} & 71.68 & 72.15 & \textbf{73.25}\\
        {\scriptsize ACE05-Evt (Argument)} & 55.37 & \textbf{57.77} & 57.27\\
        16-res & 76.73 & 77.13 & \textbf{78.13}\\
        \midrule
        Average & 74.48 & 75.44 & \textbf{75.75} \\
        \bottomrule
    \end{tabular}
\end{table}

\begin{table*}[th]
    \centering
    \caption{The correlation between schema complexity, training data size and relative improvement.}
    \label{tab:complexity-datasize-improvement}
    \begin{tabular}{cm{2.2cm}<{\centering}m{2.2cm}<{\centering}cccm{2cm}<{\centering}}
    \toprule
        \textbf{Dataset} & \textbf{Schema Complexity $C$} & \textbf{Training Data Size $S$} & \textbf{$\log (10000 \times \frac{C}{S})$} & \textbf{USM} & \textbf{RexUIE-EN} & \textbf{Relative Improvement}\\
        \midrule
        CoNLL03	&7	&14041	&0.6977	&92.79	&93.31	&0.56\%\\
        ACE04	&4	&6202	&0.8095&	87.79&	88.02&	0.26\%\\
        ACE05-Ent&	7	&7299&	0.9818	&86.98&	86.87&	-0.13\%\\
        NYT&	55	&56196	&0.9907&	94.07	&94.55&	0.51\%\\
        14-res&	3	&1266&	1.3747&	76.35	&76.36&	0.01\%\\
        14-lap	&3	&906	&1.5200&	65.46&	66.92	&2.23\%\\
        16-res&	3	&857&	1.5441	&76.73	&78.13&	1.82\%\\
        ACE05-Evt-Arg	&79&	19216	&1.6140	&55.37	&57.27&	3.43\%\\
        15-res	&3	&605	&1.6954	&68.8	&70.48&	2.44\%\\
        CoNLL04	&6	&922&	1.8134	&75.86&	76.79	&1.23\%\\
        SciERC&	115&	1861&	2.7910	&37.05	&38.16	&3.00\%\\
    \bottomrule
    \end{tabular}
\end{table*}

To prove the necessity of constraints by explicit schema instructor, we analyse the distribution of relation type versus \textit{subject type}-\textit{object type} predicted by T5-UIE as illustrated in Figure \ref{fig:explicit schema}.

We observe that illegal extractions, such as \textit{person, work for, location}, are not rare in 1-Shot, and a considerable number of subjects or objects are not properly extracted during the NER stage. For example, more than half of the predictions of the relation \textit{located in}, end with a \textit{none} type entities. 29\% of the predictions about relation \textit{live in} end with \textit{none} type entities.
Although this issue is alleviated in the 5-Shot scenario, there are still 43\% wrong predictions of \textit{located in} violated the schema. We believe that the implicit schema instructor still negatively affects the model's performance. Without the constraints on correspondence between types, the model has no idea about what types are allowed to be extracted following the current result. This behaviour obviously ignores the knowledge pre-defined in the schema.

\subsection{Ablation Study}\label{sec:ablation study}

We conducted an ablation experiment on RexUIE-EN to explore the influence of prompts isolation and rotary embedding on the model, where RexUIE-EN is not pre-trained. The results are listed in Table \ref{tab:ablation study}.

The experimental results demonstrate that removing prompts isolation leads to a decrease in the performance of RE and event trigger. The F1 score of RE decreases from 76.79\% to 73.76\%. The event trigger F1 decreases from 73.25\% to 73.10\%.
The exclusion of rotary embedding results in detrimental effects on both relation and sentiment extraction. By removing rotary embedding, the relative performance drop on RE F1 is 3.78\% and 4.35\% on 16-res.
Overall, the complete RexUIE-EN exhibits superior performance. Removing prompts isolation or rotary embedding results in a slight decline in performance, with the most significant drop observed when both are deleted. The average F1 by removing both of them is 73.83\%, which is a relative decrease of 2.53\%.

\subsection{Influence of Encoder}


DeBERTa \cite{he2021deberta} improved BERT \cite{bert} and RoBERTa \cite{liu2019roberta} models using the disentangled attention mechanism and enhanced mask decoder.
To ensure that our work was not entirely dependent on a better text encoder, we replaced DeBERTaV3-Large with RoBERTa-Large, denoted as $\text{RexUIE-EN}_{\text{RoBERTa}}$, thus maintaining the same text encoder settings as USM. We compare these two models with USM, which is also produced based on RoBERTa-Large. The comparison on the datasets is presented in Table \ref{tab:rexuie-roberta}.

$\text{RexUIE-EN}_{\text{RoBERTa}}$ achieved two best results, and RexUIE-EN achieved three best results, slightly surpassing the former. Notably, F1 of event argument by $\text{RexUIE-EN}_{\text{RoBERTa}}$ is 57.77, 2.4\% higher than USM. The average result of $\text{RexUIE-EN}_{\text{RoBERTa}}$ 0.96 higher than USM.
In summary, $\text{RexUIE-EN}_{\text{RoBERTa}}$ shows an middle performance level between USM and RexUIE-EN, demonstrating that our proposed approach does indeed improve performance, and incorporating DeBERTaV3 to RexUIE-EN further enhances this improvement.

\subsection{Insights to the Schema Complexity and Training Data Size}

Under the full-shot setting, the improvement of RexUIE-EN compared to the previous UIE models is not significant (only 1\% across 4 tasks and 14 metrics). At the same time, we have also found that for different tasks or datasets, the improvement of RexUIE-EN seems to exhibit some randomness, which may be related to several factors such as schema complexity, training data size, task type, and the extent of task exploration. Among these factors, we believe that schema complexity and training data size are more important, so we conducted a statistical analysis to better summarize the patterns. (We only consider the case of full-shot without pre-training to avoid the influence of pre-training.)

\begin{itemize}
    \item Schema complexity: Due to ESI and recursive strategies, we intuitively believe that RexUIE-EN has certain advantages in handling complex schemas. We use the number of leaf nodes in the schema to represent the complexity of the schema, noted as $C$.
    \item Training data size: We know that as the training data size increases, the differences between the performance of models will be narrow. Therefore, we believe that the performance improvement is negatively correlated with training data size. We note the training data size as $S$.
\end{itemize}
To investigate the pattern, we introduce a media variable $\log (10000 \times \frac{C}{S})$. After removing certain outliers and event-trigger datasets, we find a positive correlation between $\log (10000 \times \frac{C}{S})$ and the relative improvement in Table \ref{tab:complexity-datasize-improvement}, which supports our hypothesis.

\section{Example of Schema}\label{sec:schema case}

Schema examples for some datasets are listed in Table \ref{tab: schema example}.

\begin{table*}
    \centering
    \footnotesize
    \caption{Schema examples.}
    \label{tab: schema example}
    \begin{tabular}{m{1.5cm}|m{8cm}|m{5cm}}
    \toprule
        Dataset & Schema $\mathbf{C}$ & Example of $\mathbf{t}$\\
    \midrule
        Entity CoNLL03 & \{``person'': null, ``location'': null, ``miscellaneous'': null, ``organization'': null\} & [``person''] \\
    \midrule
        Relation CoNLL04 & \{``organization'': \{``organization in ( location )'': null\}, ``other'': null, ``location'': \{``located in ( location )'': null\}, ``people'': \{``live in ( location )'': null, ``work for ( organization )'': null, ``kill ( people )'': null\}\} & [``organization'', ``organization in ( location )'']\\
    \midrule
        Event ACE05-Evt & \{``attack'': \{``attacker'': null, ``place'': null, ``target'': null, ``instrument'': null\}, ``end position'': \{``person'': null, ``place'': null, ``entity'': null\}, ``meet'': \{``place'': null, ``entity'': null\}, ``transport'': \{``artifact'': null, ``origin'': null, ``destination'': null, ``agent'': null, ``vehicle'': null\}, ``die'': \{``victim'': null, ``instrument'': null, ``place'': null, ``agent'': null\}, ``transfer money'': \{``giver'': null, ``beneficiary'': null, ``recipient'': null\}, ``trial hearing'': \{``adjudicator'': null, ``defendant'': null, ``place'': null\}, ``charge indict'': \{``defendant'': null, ``place'': null, ``adjudicator'': null\}, ``transfer ownership'': \{``beneficiary'': null, ``artifact'': null, ``seller'': null, ``place'': null, ``buyer'': null\}, ``sentence'': \{``defendant'': null, ``place'': null, ``adjudicator'': null\}, ``extradite'': \{``person'': null, ``destination'': null, ``agent'': null\}, ``start position'': \{``place'': null, ``entity'': null, ``person'': null\}, ``start organization'': \{``organization'': null, ``agent'': null, ``place'': null\}, ``sue'': \{``defendant'': null, ``plaintiff'': null\}, ``divorce'': \{``person'': null, ``place'': null\}, ``marry'': \{``person'': null, ``place'': null\}, ``phone write'': \{``place'': null, ``entity'': null\}, ``injure'': \{``victim'': null\}, ``end organization'': \{``organization'': null\}, ``appeal'': \{``adjudicator'': null, ``plaintiff'': null\}, ``convict'': \{``defendant'': null, ``place'': null, ``adjudicator'': null\}, ``fine'': \{``entity'': null, ``adjudicator'': null\}, ``declare bankruptcy'': \{``organization'': null\}, ``demonstrate'': \{``place'': null, ``entity'': null\}, ``elect'': \{``person'': null, ``place'': null, ``entity'': null\}, ``nominate'': \{``person'': null\}, ``acquit'': \{``defendant'': null, ``adjudicator'': null\}, ``execute'': \{``agent'': null, ``person'': null, ``place'': null\}, ``release parole'': \{``person'': null\}, ``arrest jail'': \{``person'': null, ``place'': null, ``agent'': null\}, ``born'': \{``person'': null, ``place'': null\}\} & [``attack'', ``attacker'']\\
        \midrule
        Sentiment 16-res & \{``aspect'': \{``positive ( opinion )'': null, ``neutral ( opinion )'': null, ``negative ( opinion )'': null\}, ``opinion'': null\} & [``aspect'', ``positive ( opinion )'']\\
        \midrule
        COQE Camera & \{``subject'': \{``object'': \{``aspect'': \{``worse ( opionion )'': null, ``equal ( opinion )'': null, ``better ( opinion )'': null, ``different ( opinion )'': null\}, ``worse ( opionion )'': null, ``equal ( opinion )'': null, ``better ( opinion )'': null, ``different ( opinion )'': null\}, ``aspect'': \{``worse ( opionion )'': null, ``equal ( opinion )'': null, ``better ( opinion )'': null, ``different ( opinion )'': null\}, ``worse ( opionion )'': null, ``equal ( opinion )'': null, ``better ( opinion )'': null, ``different ( opinion )'': null\}, ``object'': \{``aspect'': \{``worse ( opionion )'': null, ``equal ( opinion )'': null, ``better ( opinion )'': null, ``different ( opinion )'': null\}, ``worse ( opionion )'': null, ``equal ( opinion )'': null, ``better ( opinion )'': null, ``different ( opinion )'': null\}, ``aspect'': \{``worse ( opionion )'': null, ``equal ( opinion )'': null, ``better ( opinion )'': null, ``different ( opinion )'': null\}, ``worse ( opionion )'': null, ``equal ( opinion )'': null, ``better ( opinion )'': null, ``different ( opinion )'': null\}\} & [``subject'', ``object'', ``aspect'', ``better ( opinion )'']\\
        \bottomrule
    \end{tabular}
\end{table*}

\section{Query Example}\label{sec:show case}

Some query examples are listed in Table \ref{tab:sample of query 1} and Table \ref{tab:sample of query 2}.

\begin{table*}
    \centering
    \caption{Query examples for CoNLL03, CoNLL04 and ACE05-Evt.}
    \label{tab:sample of query 1}
    \begin{tabular}{c|c|m{10cm}}
    \toprule
        Dataset & Sample Id & Query \\
    \midrule
        CoNLL03 & 0 & [CLS][P][T] location[T] miscellaneous[T] organization[T] person[Text] EU rejects German call to boycott British lamb .[SEP] \\
        \midrule
        \multirow{2}{*}{CoNLL04} & \multirow{2}{*}{0} &  [CLS][P][T] location[T] organization[T] other[T] people[Text] The self-propelled rig Avco 5 was headed to shore with 14 people aboard early Monday when it capsized about 20 miles off the Louisiana coast , near Morgan City , Lifa said.[SEP]\\
        \cline{3-3}
        & & [CLS][P] location: Morgan City[T] located in ( location )[P] location: Louisiana[T] located in ( location )[P] people: Lifa[T] kill ( people )[T] live in ( location )[T] work for ( organization )[Text] The self-propelled rig Avco 5 was headed to shore with 14 people aboard early Monday when it capsized about 20 miles off the Louisiana coast , near Morgan City , Lifa said.[SEP]\\
        \midrule
        \multirow{4}{*}{ACE05-Evt} & \multirow{2}{*}{0} & [CLS][P][T] acquit[T] appeal[T] arrest jail[T] attack[T] born[T] charge indict[T] convict[T] declare bankruptcy[T] demonstrate[T] die[T] divorce[T] elect[T] end organization[T] end position[T] execute[T] extradite[T] fine[T] injure[T] marry[T] meet[T] merge organization[T] nominate[T] pardon[T] phone write[T] release parole[T] sentence[T] start organization[T] start position[T] sue[T] transfer money[T] transfer ownership[T] transport[T] trial hearing[Text] The electricity that Enron produced was so exorbitant that the government decided it was cheaper not to buy electricity and pay Enron the mandatory fixed charges specified in the contract .[SEP]\\
        \cline{3-3}
        & & [CLS][P] transfer money: pay[T] beneficiary[T] giver[T] place[T] recipient[Text] The electricity that Enron produced was so exorbitant that the government decided it was cheaper not to buy electricity and pay Enron the mandatory fixed charges specified in the contract .[SEP]\\
        \cline{2-3}
        & \multirow{2}{*}{1} & [CLS][P][T] acquit[T] appeal[T] arrest jail[T] attack[T] born[T] charge indict[T] convict[T] declare bankruptcy[T] demonstrate[T] die[T] divorce[T] elect[T] end organization[T] end position[T] execute[T] extradite[T] fine[T] injure[T] marry[T] meet[T] merge organization[T] nominate[T] pardon[T] phone write[T] release parole[T] sentence[T] start organization[T] start position[T] sue[T] transfer money[T] transfer ownership[T] transport[T] trial hearing[Text] and he has made the point repeatedly in interview after interview that he has never claimed to speak for god , nor has he claimed that this is `` god \' s war ''[SEP]\\
        \cline{3-3}
        & & [CLS][P] attack: war[T] attacker[T] instrument[T] place[T] target[T] victim[Text] and he has made the point repeatedly in interview after interview that he has never claimed to speak for god , nor has he claimed that this is `` god \' s war ''[SEP]\\
    \bottomrule
    \end{tabular}
    
\end{table*}

\begin{table*}
    \centering
    \caption{Query examples for 16-res and COQE-Camera.}
    \label{tab:sample of query 2}
    \begin{tabular}{c|c|m{10cm}}
    \toprule
        Dataset & Sample Id & Query \\
    \midrule
        \multirow{4}{*}{16-res} & \multirow{2}{*}{0} & [CLS][P][T] aspect[T] opinion[Text] Judging from previous posts this used to be a good place , but not any longer .[SEP] \\
        \cline{3-3}
        & & [CLS][P] aspect: place[T] negative ( opinion )[T] neutral ( opinion )[T] positive ( opinion )[Text] Judging from previous posts this used to be a good place , but not any longer .[SEP]\\
        \cline{2-3}
        & \multirow{2}{*}{1} & [CLS][P][T] aspect[T] opinion[Text] The food was lousy - too sweet or too salty and the portions tiny .[SEP]\\
        \cline{3-3}
        & & [CLS][P] aspect: portions[T] negative ( opinion )[T] neutral ( opinion )[T] positive ( opinion )[P] aspect: food[T] negative ( opinion )[T] neutral ( opinion )[T] positive ( opinion )[Text] The food was lousy - too sweet or too salty and the portions tiny .[SEP]\\
    \midrule
        \multirow{3}{*}{COQE-Camera} & \multirow{3}{*}{0} & [CLS][P][T] aspect[T] better ( opinion )[T] different ( opinion )[T] equal (opinion )[T] object[T] subject[T] worse ( opionion )[Text] Also , both the Nikon D50 and D70S will provide sharper pictures with better color saturation and contrast right out of the camera .[SEP]\\
        \cline{3-3}
        && [CLS][P] subject: Nikon D50[T] aspect[T] better ( opinion )[T] different ( opinion )[T] equal (opinion )[T] object[T] worse ( opionion )[P] subject: D70S[T] aspect[T] better ( opinion )[T] different ( opinion )[T] equal (opinion )[T] object[T] worse ( opionion )[Text] Also , both the Nikon D50 and D70S will provide sharper pictures with better color saturation and contrast right out of the camera .[SEP]\\
        \cline{3-3}
        &&[CLS][P] subject: D70S,aspect: pictures[T] better ( opinion )[T] different ( opinion )[T] equal (opinion )[T] worse ( opionion )[P] subject: D70S,aspect: color saturation[T] better ( opinion )[T] different ( opinion )[T] equal (opinion )[T] worse ( opionion )[P] subject: Nikon D50,aspect: pictures[T] better ( opinion )[T] different ( opinion )[T] equal (opinion )[T] worse ( opionion )[P] subject: Nikon D50,aspect: contrast[T] better ( opinion )[T] different ( opinion )[T] equal (opinion )[T] worse ( opionion )[P] subject: Nikon D50,aspect: color saturation[T] better ( opinion )[T] different ( opinion )[T] equal (opinion )[T] worse ( opionion )[P] subject: D70S,aspect: contrast[T] better ( opinion )[T] different ( opinion )[T] equal (opinion )[T] worse ( opionion )[Text] Also , both the Nikon D50 and D70S will provide sharper pictures with better color saturation and contrast right out of the camera .[SEP]\\
    \bottomrule
    \end{tabular}
    
\end{table*}

}

 




\vfill

\end{document}